\documentclass{article}

\usepackage{arxiv}

\usepackage[utf8]{inputenc} 
\usepackage[T1]{fontenc}    
\usepackage{hyperref}       
\usepackage{url}            
\usepackage{booktabs}       
\usepackage{amsfonts}       
\usepackage{nicefrac}       
\usepackage{microtype}      
\usepackage{lipsum}         
\usepackage{graphicx}       
\usepackage{natbib}         
\usepackage{doi}            
\usepackage{amsmath}        
\usepackage{amssymb}        
\usepackage{latexsym}       
\usepackage{mathtools}      
\usepackage{multirow}       
\usepackage{caption}        
\usepackage{stfloats}       
\usepackage[table]{xcolor}  
\usepackage{float}          
\usepackage{tabularx}       
\usepackage{array}          
\usepackage{authblk}  
\usepackage{hyperref} 

\DeclarePairedDelimiter\abs{\lvert}{\rvert} 

\usepackage{etoolbox}
\usepackage{pgf} 
\definecolor{high}{HTML}{f8696b}  
\definecolor{mid}{HTML}{ffffff}  
\definecolor{low}{HTML}{63be7b}  

\definecolor{higheta}{HTML}{588bca}  
\definecolor{loweta}{HTML}{ffffff}  

\newcommand*{\opacity}{80}
\newcommand*{\minval}{0.0}
\newcommand*{\midval}{0.1}
\newcommand*{\maxval}{0.2}

\newcommand*{\minvaleta}{0.05}
\newcommand*{\maxvaleta}{0.4}

\newcommand{\gradient}[1]{
    \ifdimcomp{#1pt}{>}{\maxval pt}{
        \cellcolor{high!\opacity} #1
    }{
        \ifdimcomp{#1pt}{<}{\minval pt}{
            \cellcolor{low!\opacity} #1
        }{
            \pgfmathparse{int(round(100 * ((#1 - \minval) / (\midval - \minval))))}
            \xdef\tempa{\pgfmathresult}
            \pgfmathparse{int(round(100 * ((#1 - \midval) / (\maxval - \midval))))}
            \xdef\tempb{\pgfmathresult}
            \ifdimcomp{#1pt}{<}{\midval pt}{
                \cellcolor{mid!\tempa!low!\opacity} #1
            }{
                \cellcolor{high!\tempb!mid!\opacity} #1
            }
        }
    }
}

\newcommand{\gradienteta}[1]{
    \ifdimcomp{#1pt}{>}{\maxvaleta pt}{
        \cellcolor{higheta!\opacity} #1
    }{
        \ifdimcomp{#1pt}{<}{\minvaleta pt}{
            \cellcolor{loweta!\opacity} #1
        }{
            \pgfmathparse{int(round(100 * ((#1 - \minvaleta) / (\maxvaleta - \minvaleta))))}
            \xdef\tempa{\pgfmathresult}
            \cellcolor{higheta!\tempa!loweta!\opacity} #1
        }
    }
}

\newcommand{\gradientetados}[1]{
    \ifdimcomp{#1pt}{>}{\maxvaleta pt}{
        \cellcolor{high!\opacity} #1
    }{
        \ifdimcomp{#1pt}{<}{\minvaleta pt}{
            \cellcolor{loweta!\opacity} #1
        }{
            \pgfmathparse{int(round(100 * ((#1 - \minvaleta) / (\maxvaleta - \minvaleta))))}
            \xdef\tempa{\pgfmathresult}
            \cellcolor{high!\tempa!loweta!\opacity} #1
        }
    }
}

\title{A Computational Pipeline for Advanced Analysis of\\4D Flow MRI in the Left Atrium}

\date{} 					

\definecolor{affilcolor}{RGB}{128, 0, 0}  

\author[1]{\raggedright Xabier Morales\thanks{Corresponding author: xtaltec@gmail.com}}
\author[2,3]{Ayah Elsayed}
\author[2]{Debbie Zhao}
\author[8]{Filip Loncaric}
\author[1]{Ainhoa Aguado}
\author[1]{Mireia Masias}
\author[2]{Gina Quill}
\author[5,6]{Marc Ramos}
\author[5,6]{Ada Doltra}
\author[5,6]{Ana Garcia}
\author[5,6]{Marta Sitges}
\author[10,11]{David Marlevi}
\author[2]{Alistair Young}
\author[2,4]{Martyn Nash}
\author[1,5]{Bart Bijnens}
\author[1]{Oscar Camara}

\affil[1]{\raggedright PhySense, Department of Information and Communication Technologies, Universitat Pompeu Fabra, Barcelona, Spain}
\affil[2]{\raggedright Auckland Bioengineering Institute, University of Auckland, Auckland, New Zealand}
\affil[3]{\raggedright Faculty of Health and Environmental Sciences, Auckland University of Technology, Auckland, New Zealand}
\affil[4]{\raggedright Department of Engineering Science and Biomedical Engineering, University of Auckland, Auckland, New Zealand}
\affil[5]{\raggedright Cardiovascular Institute, Hospital Clínic, Universitat de Barcelona, Barcelona, Spain}
\affil[6]{\raggedright Institut d'investigacions biomèdiques august pi i sunyer (IDIBAPS), Barcelona, Spain}
\affil[7]{\raggedright Institució Catalana de Recerca i Estudis Avançats, (ICREA), Barcelona, Spain}
\affil[8]{\raggedright University Hospital Centre Zagreb, Zagreb, Croatia}
\affil[9]{\raggedright School of Biomedical Engineering \& Imaging Sciences, King’s College London, London, UK}
\affil[10]{\raggedright Institute for Medical Engineering and Science, Massachusetts Institute of Technology, Cambridge, MA, USA}
\affil[11]{\raggedright Karolinska Institutet, Department of Molecular Medicine and Surgery, Stockholm, Sweden}

\renewcommand{\undertitle}{}
\renewcommand{\shorttitle}{\raggedright Computational Pipeline for 4D Flow MRI Analysis in the Left Atrium}

\hypersetup{
    pdftitle={A Computational Pipeline for Advanced Analysis of 4D Flow MRI in the Left Atrium},
    pdfsubject={Medical Imaging, MRI, Computational Pipeline},
    pdfauthor={Xabier Morales, Ayah Elsayed, Debbie Zhao, Filip Loncaric, Ainhoa Aguado, Mireia Masias, Gina Quill, Marc Ramos, Ada Doltra, Ana Garcia, Marta Sitges, David Marlevi, Alistair Young, Martyn Nash, Bart Bijnens, Oscar Camara},
    pdfkeywords={4D Flow MRI, Left Atrium, Medical Imaging, Computational Analysis, Cardiovascular Research},
}

\begin{document}
\maketitle

\begin{abstract}

The left atrium (LA) plays a pivotal role in modulating left ventricular filling, but our comprehension of its hemodynamics is significantly limited by the constraints of conventional ultrasound analysis. 4D flow magnetic resonance imaging (4D Flow MRI) holds promise for enhancing our understanding of atrial hemodynamics. However, the low velocities within the LA and the limited spatial resolution of 4D Flow MRI make analyzing this chamber challenging. Furthermore, the absence of dedicated computational frameworks, combined with diverse acquisition protocols and vendors, complicates gathering large cohorts for studying the prognostic value of hemodynamic parameters provided by 4D Flow MRI. In this study, we introduce the first open-source computational framework tailored for the analysis of 4D Flow MRI in the LA, enabling comprehensive qualitative and quantitative analysis of advanced hemodynamic parameters. Our framework proves robust to data from different centers of varying quality, producing high-accuracy automated segmentations (Dice $>$ 0.9 and Hausdorff 95 $<$ 3 mm), even with limited training data. Additionally, we conducted the first comprehensive assessment of energy, vorticity, and pressure parameters in the LA across a spectrum of disorders to investigate their potential as prognostic biomarkers.

\end{abstract}


\section{Introduction}

The left atrium (LA) constitutes one of the four chambers of the heart, playing a pivotal role in modulating left ventricular filling. Although the left ventricle (LV) was long held as the sole determinant of cardiac health prognosis, alterations in LA structure, function, and hemodynamics are now recognized as pivotal factors in various cardiovascular disorders, including atrial fibrillation (AF), heart failure, and ischemic and valvular heart disease \cite{Fyrenius2001, inciardi2019left}. For instance, AF-related hemodynamic disruption of the LA can lead to thrombus formation, increasing the risk of cerebrovascular thromboembolic events \cite{Fang2022}. Additionally, abnormalities in LA hemodynamics often precede the clinical manifestation of LV diastolic dysfunction (LVDD), which remains difficult to characterize non-invasively \cite{MacNamara2021,Ashkir2022}.

Despite its significant role, the analysis of LA hemodynamics has garnered considerably less attention than more prominent structures like the LV or the aorta \cite{Marino2019}. This issue is compounded by the fact that spectral Doppler, considered the gold standard for flow assessment in clinical routine, is ill-suited to fully characterize the intricate 3D hemodynamics present in the LA \cite{Vedula2015}, as it can only provide velocity in one dimension along the beam line. Moreover, when performed through transthoracic echocardiography (TTE), precise measurement of the pulmonary veins (PV), relevant for the diagnosis of LVDD, becomes highly challenging due to their location in the far field \cite{Huang2008}.  

In contrast, 3D time-resolved phase-contrast magnetic resonance imaging (MRI), often referred to as 4D Flow MRI, has emerged as a promising tool for conducting comprehensive in vivo hemodynamic studies. 4D Flow MRI enables the acquisition of time-resolved 3D blood velocity with full volumetric coverage, enabling retrospective quantification of flow in a single non-invasive examination \cite{Bissell2023}. This capability opens up the potential for scrutinizing the physiology of inherently three-dimensional structures, such as LA vortices, akin to previous studies conducted on the LV \cite{Kruter2020}, or the concurrent analysis of all four PV, seldom available with conventional TTE analyses \cite{Blume2011}. 

However, the low spatial resolution due to the need to maintain reasonable acquisition times, combined with the low velocities present, has severely hampered the use of 4D Flow MRI in the LA. Moreover, despite the availability of both commercial and free software for 4D Flow MRI analysis \cite{Khler2019}, existing pipelines primarily focus on aortic or ventricular flow, which largely precludes the analysis of LA hemodynamics for those not involved in research. The issue is exacerbated by the heterogeneous nature of 4D flow MRI acquisition, characterized by varying spatiotemporal resolution and noise levels among different MRI vendors, magnetic field strengths, acquisition protocols, and contrast utilization. This makes it challenging to standardize the analysis pipelines necessary to amass sufficiently large cohorts by gathering data from different centers. Consequently, the number of published studies focusing on LA hemodynamics using 4D Flow MRI remains limited \cite{Fluckiger2013,Markl2016,Gaeta2018,Garcia2019,Demirkiran2021,Spartera2021,Nallamothu2024}, with few of them fully exploring the plethora of novel hemodynamic parameters as potential prognostic biomarkers.

In light of these challenges, this study aims to introduce a semi-automated computational pipeline specifically designed for the analysis of 4D Flow MRI in the LA. This pipeline is robust enough to handle multiple center data of varying quality, allowing for the comprehensive visualization of LA hemodynamics and accurate quantification of advanced 4D Flow MRI parameters. We have also developed a method for direct comparison with ultrasound pulsed-wave (PW) Doppler velocimetry. This comparison serves to highlight the advantages and potential applications of 4D Flow MRI in a clinical setting.

Furthermore, to make this tool accessible to the wider clinical community, we have emphasized the use of open-source code and software in its development. By doing so, we hope to encourage further research and innovation in LA hemodynamics, ultimately leading to improved understanding and treatment of related conditions.

The main contributions of the present article are:

\begin{itemize}
    \item We introduce the first 4D Flow MRI analysis pipeline tailored to the LA, based entirely on open-source software and Python.
    
    \item  We employ deep learning to automate left atrium segmentation, achieving high accuracy with minimal data. The pipeline readily adapts to new datasets by incorporating a few high-quality examples into the training dataset.

    \item We showcase the feasibility of generating PW Doppler-like velocity spectrograms from 4D Flow MRI, and we explore their potential applications.
    
    \item  We conduct the first comprehensive analysis of novel 4D Flow MRI biomarkers in the LA across various disorders exploring their prognostic potential.
    
\end{itemize}

\section{Material and Methods}

\begin{figure}[]
\centering
\includegraphics[width=\textwidth]{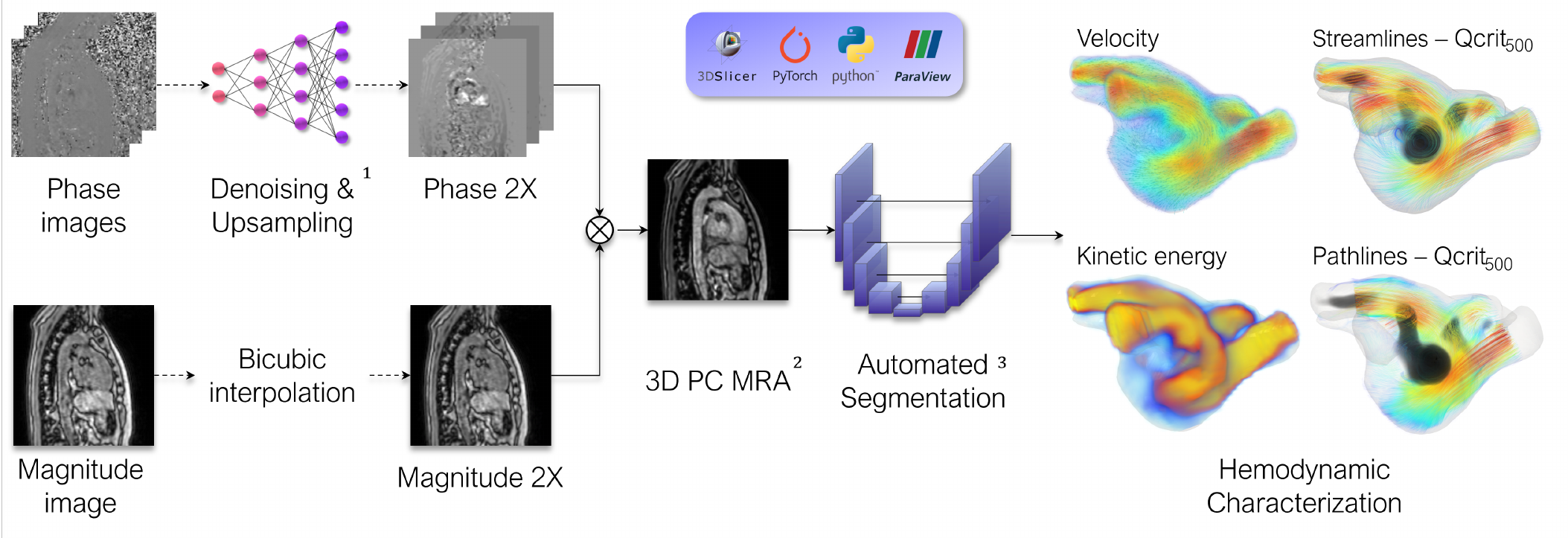}
\caption{Overview of the computational framework for the advanced analysis of 4D flow magnetic resonance imaging of the left atrium. Only data from lower magnetic field strength acquisitions undergoes denoising and upsampling [1]\cite{4dflownet}. Next, a 3D PC-MRA is computed [2]\cite{Bustamante2017}, followed by automatic segmentation [3]\cite{nnunet}. The resulting segmentation mask facilitates the isolation of the structure of interest for subsequent quantitative and qualitative hemodynamic characterization. The entire pipeline relies exclusively on open-source software and code, ensuring accessibility and reproducibility. PC-MRA: Phase-contrast Magnetic Resonance Angiogram; Q-crit$_{500}$: Ratio of Q-criterion $>$ 500 s$^{-2}$.}
\label{fig:General Pipeline}
\end{figure}

Figure \ref{fig:General Pipeline} provides a high-level overview of the framework. First, lower-resolution acquisitions undergo denoising and upsampling. Afterward, the phase and magnitude images are combined to generate a 3D phase-contrast magnetic resonance angiogram (3D PC-MRA) that is automatically segmented. The resulting segmentation is then used to mask the flow data, which is transformed into a format suitable for subsequent qualitative and quantitative analysis.

\subsection{Population and Data Acquisition}

To showcase the versatility of the pipeline, we analyzed a dataset comprising 109 subjects, incorporating data from two centers. The Centre of Advanced MRI (CAMRI) at the University of Auckland, New Zealand, recruited 39 healthy participants. Additionally, we included 29 patients with hypertrophic cardiomyopathy (HCM) and varying degrees of LVDD, and 41 hypertensive patients from the Department of Cardiology, Hospital Clínic de Barcelona, Spain. Ethical approval was granted by the respective research ethics committees at each participating center, and written informed consent was obtained from all study participants.

The CAMRI data were acquired using a Siemens Magnetom 1.5~T Avanto Fit (Siemens Healthcare, Erlangen, Germany) scanner with the following acquisition parameters: axial slab orientation, retrospective cardiac gating, free breathing, 30-channel coil array, velocity encoding = 150 cm/s, flip angle = 7º, echo time = 2.3 ms, repetition time = 38.8 ms, voxel size = 2.4 x 2.375 x 2.375 mm acquired and reconstructed, a field of view of 125 mm x 285 mm x 380 mm$^{3}$, acquisition matrix 52 x 110 x 160, 20 phases reconstructed, acceleration factor of 3 with an average scan time of 8 minutes and no contrast used. The remaining data were obtained with a SIGNA Architect 3.0T MRI scanner (General Electric Medical Systems) featuring an axial slab orientation, retrospective cardiac gating, respiration compensation 10\% of k-space, small anterior 16-channel and spine posterior 40-channel coil array, velocity encoding = 160 cm/s, a flip angle of 15º, an echo time of 2.216 ms, a repetition time of 4.172 ms, a voxel size of 2.2 mm x 2.2 mm x 2.2 mm acquired and 1.4063 mm x 1.40635 mm x 1.1 mm reconstructed, a field of view of 360 x 360 x 176 mm$^{3}$, acquisition matrix 160 x 160 x 90, 30 phases reconstructed, HyperKat acceleration with a factor of 8, an average scan time of 10 minutes and the use of contrast (Gadoterate meglumine 0.1 mmol/kg). Hereafter, we will denote the datasets from the Siemens and General Electric scanners as datasets 1.5~T and 3~T, respectively. Processing of the raw phase data, including background phase offset correction and velocity anti-aliasing, was done at the acquisition centers with \href{https://www.arterys.com/}{Arterys}. 

In addition, 2D balanced steady-state free precession (bSSFP) cine MRI images of the two-chamber and four-chamber views were acquired for a total of 42 patients, including 21 patients from the 1.5~T cohort and 21 from the 3~T cohort. Furthermore, 88 patients (23 controls, 27 HCM, and 38 hypertensive) underwent PW Doppler velocimetry via TTE in the apical four-chamber view. Mitral velocities were recorded at the mitral leaflets, and right superior (RS) PV velocities were recorded whenever feasible. Subsequently, all patients underwent echocardiographic LVDD grading according to the 2016 ASE/EACVI guidelines \cite{Nagueh2016}.

\subsection{Data denoising and upsampling}

Diverse magnetic field strengths, acquisition protocols, and vendors result in 4D Flow MRI data of differing spatial resolution and quality across centers. This can have a substantial impact on the geometry and quantitative measurements of velocity \cite{Cherry2022}. As it is often necessary to pool data from several centers to gather a sufficiently large dataset for clinical studies, it is imperative to minimize the differences in spatial resolution and image quality to reduce measurement bias.

In this regard, we applied the 4DFlowNet super-resolution network \cite{4dflownet} to simultaneously denoise and upsample the phase images by a factor of two. As the neural network was only trained to upsample the phase image, the resolution of the magnitude image was doubled through bicubic interpolation. Only data from the 1.5~T scanner was upsampled using the available pre-trained model, resulting in a final spatial resolution of 1.2 mm x 1.1875 mm x 1.1875 mm, closely matching the 3~T dataset.

\subsection{Segmentation} \label{sec:Segmentation}

Precise segmentation of the region of interest significantly enhances visual qualitative analysis and enables the quantification of clinically relevant parameters. Nevertheless, segmentation remains a challenging aspect in 4D Flow MRI analysis, particularly in the LA, primarily due to its low spatial resolution \cite{MarinCastrillon2023}. The time-consuming nature of manual segmentation further hinders the broader adoption of 4D Flow MRI in clinical settings \cite{GarridoOliver2022}.

\subsubsection{PC-MRA} \label{sec: PC-MRA}

4D Flow MRI segmentation commonly involves generating a phase-contrast magnetic resonance angiogram (PC-MRA). In the resulting image, phase data yields higher intensities in areas of high blood flow, while the magnitude image adds morphological information and mitigates noise in regions with low signal, such as the lungs. We utilized the equation proposed by \cite{Bustamante2017}, which incorporates a correction factor $\gamma < 1$ to enhance low-velocity areas. While the original article considered a time-resolved PC-MRA, segmentation of the LA in timesteps of low velocities proved infeasible. Consequently, we adopted a time-averaged version of the equation introduced by \cite{Bustamante2017}:

\begin{equation}\label{eq:LA-PC-MRA}
    \textrm{3D PC-MRA} = \frac{1}{N} \sum_{t=1}^{N} M(t) \cdot ({V_x}^{2}(t)+{V_y}^{2}(t)+{V_z}^{2}(t))^{\gamma}\,,
\end{equation}

\noindent where $M(t)$ is the signal magnitude and ${V_x}$, ${V_y}$, and ${V_z}$ are the three components of the velocity. After qualitative visual testing, we selected $\gamma=0.4$ for the LA. The resulting PC-MRA from the different values of $\gamma$ are provided in the Appendix alongside examples of both good and poor-quality single-timestep PC-MRA.

\begin{figure}[ht!]
\centering
\includegraphics[width=\textwidth]{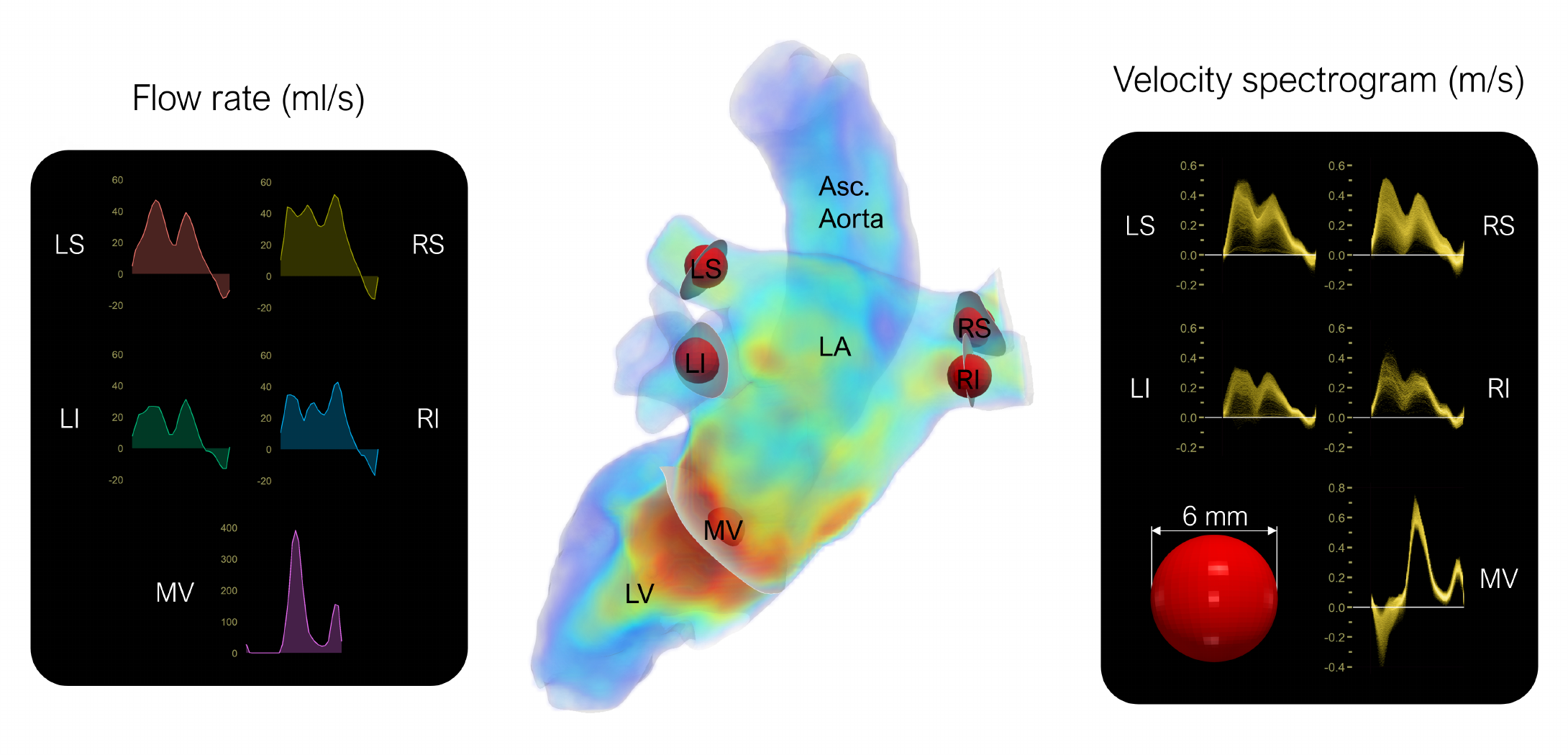}
\caption{At the center, a volumetric rendering of the 4D flow magnetic resonance imaging velocity field in the left heart is shown, encompassing the left atrium, left ventricle, and ascending aorta. Sample volumes analogous to those placed during pulsed-wave Doppler were approximated by spheres with a 6mm diameter, shown in red, displayed alongside the vessel cross-sections, depicted as gray planes. The corresponding flow rates ($ml/s$) and velocity spectrograms ($m/s)$ for the mitral valve and the four pulmonary veins are shown left and right, respectively.  LA: Left atrium, LV: Left ventricle, Asc. Aorta: Ascending aorta, MV: Mitral valve, RS: Right superior, LS: Left superior, RI: Right inferior, LI: Left inferior.}
\label{fig:Spectrogram}
\end{figure}

\subsubsection{Automatic segmentation of the left atrium} 

Besides being time-consuming, manual segmentation entails a high degree of dependence on the annotator's experience and is prone to intra- and inter-observer variability. Consequently, the segmentation was automated using the nnU-Net \cite{nnunet}, which has already proven effective in segmenting aortic 3D PC-MRA \cite{GarridoOliver2022}. The nnU-Net is a self-adaptive framework that automates the selection of optimal preprocessing, data augmentation, network architecture, and postprocessing strategies. This ensures maximum accuracy with minimal user input for each segmentation task. Training was completed on an Nvidia RTX A6000 GPU. The training dataset was manually segmented in \href{https://www.slicer.org/}{Slicer3D} by an experienced annotator. To test the capabilities of the network on 4D flow MRI data, two separate experiments were conducted.

First, we tried to determine the minimum number of cases needed to effectively train the network on a single dataset. For this purpose, we separated 60 cases from the 3~T scanner and performed successive runs, adding 10 more cases to the training dataset in each iteration. In the second experiment, we sought to find the minimum number of ground truth cases required to generalize to a new dataset. We used the 60 cases from the 3~T scanner as the starting point and successively added 5 cases from the 1.5~T dataset to the training dataset in each run. Ten cases from each dataset were allocated for testing. The performance of each model was reported in terms of the Dice score and the 95th percentile of the Hausdorff distance (HD95). The nnU-Net was trained for 1000 epochs, taking around 70 hours to complete a single fold on average. Despite the primary focus of the pipeline being on the LA, segmentation of the LV and ascending aorta was also incorporated in separate networks for visualization and computing the pressure gradients.

\subsection{Quantitative and qualitative analysis}

To extract meaningful knowledge from the high-dimensional 4D Flow MRI data, it is crucial to generate a comprehensive set of descriptive and easily understandable quantitative and qualitative metrics. In the following section, we include a detailed description of all the quantitative indices and visualizations included in our framework. We then perform a comprehensive comparison of these parameters in a diverse cohort of patients to establish their potential prognostic significance in the LA, serving as the groundwork for future studies.

\subsubsection{Segmentation post-processing} \label{sec:postpro}

After segmentation, the region of interest is isolated by zeroing all velocities outside the segmentation mask. The data is then converted to VTK image format (.vti) for use in \href{https://www.paraview.org/}{ParaView}. The segmentation mask is also converted to a triangular surface mesh (.stl) and a tetrahedral volumetric mesh (.vtk) for volume and flow rate computations. The whole process is semi-automated using Python code and ParaView state files. User input is limited to the manual placement of the spheres on the structures to be measured, as shown in Figure \ref{fig:Spectrogram}, which includes the four PVs and the MV. These spheres serve as both sample volumes for velocity spectrograms and as the origin for vessel cross-sections. The normal vector of the cross-section plane is determined by averaging the flow direction during the five timesteps with the highest velocity magnitude across the cardiac cycle. 

\subsubsection{4D Flow MRI-derived velocity spectrometry}

Ultrasound is widely regarded as the gold standard for cardiac flow analysis due to its affordability and availability. In particular, PW Doppler is essential for the non-invasive assessment of LVDD \cite{Nagueh2016}. To address the limitations of PW Doppler in imaging pulmonary flow, we aimed to derive comparable velocity spectrograms using 4D flow MRI data. To this end, idealized spherical sample volumes with a diameter of 6 mm were retrospectively placed on the masked velocity field within ParaView, as illustrated in Figure \ref{fig:Spectrogram}. Data points within each sphere were projected along the normal vector of the vessel cross-section, emulating the one-dimensional encoding of PW Doppler. This vector was determined by averaging the velocity vectors for the five timesteps with the highest velocity magnitude within the sphere. Finally, the spectrograms for each structure were generated by plotting the velocity density per timestep on the x-axis.

\subsubsection{Left atrial volume}

The indexed LA volume (LAV$_{i}$) is a robust prognostic marker of poor cardiovascular outcomes \cite{Khan2019} and an indicator of the severity and chronicity of LVDD \cite{Thomas2023}. Segmentation of the 4D Flow MRI-derived PC-MRA provides a volume estimate without the geometric assumptions of 2D imaging approaches \cite{MorAvi2012}. To evaluate the accuracy of 4D Flow MRI derived LAV${i}$ measurements, we conducted a direct comparison with the biplane disk summation method from 2D cine MRI \cite{Nacif2012}.

\subsubsection{Hemodynamic indices}

While research into the utility of 4D Flow MRI-derived parameters in the LA is gaining traction \cite{Crandon2017,Ashkir2022}, there are still considerable gaps in our understanding. To help overcome these uncertainties in future studies, our framework attempts to incorporate as many of these parameters as possible, as listed in the following:\\

\noindent\textbf{Flow rate:} This is the volume of blood passing through a given area per unit of time. In our framework, the flow rate is measured directly from the vessel cross-sections generated in ParaView. Unlike velocity, it provides information about the movement of blood volume between the different chambers without being influenced by variations in vessel cross-section.\\

\noindent\textbf{Kinetic energy:} This is the energy that blood possesses due to its motion. Changes in kinetic energy within the LA can provide insights into the atrial function and the efficiency of blood flow \cite{Gupta2021,Ashkir2022}:

\begin{equation}    
KE = \frac{1}{2} m v^2 = \frac{\rho V_{tot}}{2}\sum_{i=1}^{N} v_{i}^2 \,,
\end{equation}

\noindent where $m$ is the mass, computed as the product of the density $\rho$ and the total volume $V_{tot}$ of the region of interest, multiplied by the summation of the velocity magnitude $v_{i}$ in each voxel. \\

\noindent\textbf{Viscous energy loss:} This is the energy lost due to the frictional forces (or viscosity) within the blood flow. It can help in understanding the efficiency of blood flow within the LA \cite{Ashkir2022}. The viscous dissipation, $\phi_v$, can be computed using the Navier-Stokes energy equations \cite{Barker2013}:

\begin{equation}
\Phi_v = \frac{1}{2} \sum_{i} \sum_{j} 
\left[ \left( \frac{\partial v_i}{\partial x_j} + \frac{\partial v_j}{\partial x_i} \right) - \frac{2}{3} \left( \nabla \cdot \mathbf{v} \right)\delta_{ij} \right]
\label{eq:phi}
\end{equation}

\begin{equation}
\delta_{ij} = 
\begin{cases} 
1 & \text{for  } i = j \\ 
0 & \text{for  } i \neq j \,,
\label{eq:delta}
\end{cases}
\end{equation} 

\noindent where the first term (Equation \ref{eq:phi}) represents the symmetric part of the velocity gradient tensor and the second the divergence of the velocity field. Finally, the temporal variation of the viscous energy loss ($EL_t$) can be computed as follows:

\begin{equation}
EL_t = \mu \sum_{i=1}^{N} \Phi_v \text{Vol}_i \,,
\end{equation}

\noindent where $\mu=$0.0035 Pa $\cdot$ s is the viscosity of the blood, Vol$_i$ is the voxel volume, and $N$ is the number of voxels. \\

\noindent\textbf{Vorticity:} This is a measure of the rotation of the velocity field. Vortical structures are known to be present in the LA \cite{Fyrenius2001}, and have already been proven to be a relevant marker of LVDD in the ventricle \cite{Schfer2017}. The curl, or vorticity, is defined by:

\begin{equation} 
\boldsymbol{\omega} = \nabla \times \mathbf{v}\,,
\end{equation}

\noindent where $\nabla$ is the curl operator and $\mathbf{v}$ is the velocity field. Since the equation produces a vector field, it is commonplace to compute its magnitude, $\abs{\omega}$, for better interpretability. Additionally, we normalized $\abs{\omega}$ by the LA volume ($\abs{\omega_{LA}}$) to mitigate the bias from a dilated LA. However, $\abs{\omega}$ solely serves as a measure of the fluid's rotation rate and may be problematic in regions dominated by shear flow \cite{Gnther2018}. For this reason, we have incorporated the Q-criterion \cite{hunt1988eddies}, a parameter explicitly engineered for the detection of vortex cores:

\begin{equation}
Q = \frac{1}{2} \left( \left|\left| \boldsymbol{\Omega}\right|\right|^2 - 
\left|\left| \boldsymbol{S}\right|\right|^2\right),
\end{equation}

\noindent where $\boldsymbol{\Omega}$ denotes the vorticity tensor and $\mathbf{S}$ is the strain-rate tensor. Unlike vorticity, the Q-criterion yields a scalar field, where positive values indicate regions dominated by rotational motion, while negative values mark areas where viscous forces dominate.

Various thresholds can be selected for the Q-criterion to visualize vortical structures at varying scales. A threshold of Q-criterion $>$ 500~s$^{-2}$ was quantitatively determined based on visualizations in Figure~\ref{fig:Pathlines}. To normalize by the LA volume, we calculated the ratio of all voxels within the LA with Q-criterion $>$ 500 s$^{-2}$ (Q-crit$_{500}$). \\

\noindent\textbf{Relative pressure $\Delta P$:} 
Pressure gradients describe the underlying functional driver of blood flow, and as such, the evaluation of local pressure gradient abnormalities is an integral part of several guidelines for gauging cardiac function. Since 4D Flow MRI provides a complete description of the velocity field, pressure gradients can be derived without the constraints and assumptions required in other non-invasive modalities \cite{Marlevi2021}. Amidst the plethora of available methods, we opted for the virtual work-energy relative pressure (vWERP), which utilizes a work-energy formulation of the Navier-Stokes equation to extract pressure gradients through arbitrary vascular \cite{Marlevi2019} and cardiac structures \cite{Marlevi2021}. Although still in an exploratory phase, atrioventricular pressure gradients have been postulated as a potential marker for left heart function \cite{Marlevi2021,Vos2023}.\\

\subsubsection{Statistical analysis}

To complement the visualizations of the temporal evolution of each hemodynamic metric, a comprehensive statistical analysis of the most relevant maxima and minima was conducted to better assess their significance. In line with the longstanding goal of identifying age-independent biomarkers for non-invasive LVDD assessment, an analysis of covariance (ANCOVA) was performed to determine significant differences between cohorts while controlling for age. Mean values, along with the statistical significance (P-value) and strength of the association (effect size, $\eta^2$) for each cohort, are provided in the Appendix. If intergroup differences were found to be statistically significant ($\alpha<0.05$), a post hoc analysis was conducted using a pairwise Tukey test and Cohen's D to quantify the effect size. A Benjamini-Hochberg correction for multiple comparisons was also applied in the post hoc tests.

\subsubsection{Visualization}

The interpretation of intricate flow patterns and phenomena within the heart chambers can pose a considerable challenge. To leverage the wealth of information provided by 4D Flow MRI data, it is imperative to develop appropriate visualizations. Once the case is segmented and the spheres have been placed in the structures of interest, our pipeline automatically generates several interactive visualizations in ParaView, including volumetric renderings, vector maps, streamlines, and pathlines.

As an example of its potential clinical utility, we used one of the automatically generated visualizations to study the evolution of vortical structures in the LA, as illustrated in Figure \ref{fig:Pathlines}. This 3D representation combines a pathline visualization of the velocity and a volumetric rendering of the Q-criterion. The pathline visualization provides a detailed portrayal of flow patterns and velocities within the LA. Particles are emitted from the PV at each time step, tracing the path of each particle for the preceding 6 time steps, color-coded by velocity. The volumetric renderings of the Q-criterion approximate the location of the vortex cores. A median filter with an isotropic 3-voxel kernel was applied to eliminate noise and accentuate the largest vortex cores.

Lastly, all the visualizations generated in our pipeline can be easily exported as pre-rendered animations through ParaView. It can then be easily uploaded as a .zip file to web-based visualization platforms such as \href{https://kitware.github.io/glance/app/}{ParaView Glance} which enables easy interaction with the 3D scene without requiring any previous experience in handling visualization software. 

\section{Results}

\subsection{Population characteristics}

\begin{table}[H]
\centering
\vspace{-1em}  
\caption{Study population of the quantitative analysis: Controls; HCM: Hypertrophic cardiomyopathy with no left ventricular diastolic dysfunction (LVDD); G1: HCM with grade I LVDD; G2: HCM with grade II LVDD; G2 - SAM: HCM with grade II LVDD and systolic anterior motion of the mitral valve; Hypertensive. }

\vspace{0.5em}  

\label{table:Population}
\resizebox{0.8\textwidth}{!}{%
\begin{tabular}{ccccccc} 
\hline
  & Controls & HCM  & G1    & G2    & G2 - SAM & Hypertensive \\ \hline
N  = 68          & 19       & 9    & 5     & 6     & 4        & 25           \\
Age (years)        & 36.3 ± 2.6 & 42.9 ± 5.1  & 58.4 ± 7.2   & 52.1 ± 7.01  & 66.8 ± 3.7   & 54.4 ± 1.1 \\
Sex (M\%)        & 65     & 70 & 20 & 50  & 75     & 56        \\
BSA(Dubois) ($m^2$) & 1.93 ± 0.06  & 1.92 ± 0.04  & 2.05 ± 0.07    & 1.91 ± 0.13   & 1.78 ± 0.11    & 1.93 ± 0.04  \\
Diabetes (\%)    & 0      & 10 & 20  & 16.67 & 0      & 0          \\
Hypertension (\%) & 0      & 50 & 80  & 66.67 & 50    & 84.38        \\
SAM (M\%)         & 0        & 0    & 0     & 0     & 100      & 0            \\
LAV (ml)        & 60.4 ± 3.2 & 80.7 ± 7.5 & 116.9 ± 15.9 & 103.1 ± 24.1 & 150.5 ± 23.9 & 97.9 ± 4.6 \\
LVEDV (ml)      & 86.1 ± 5.3 & 74.5 ± 7.7   & 99.1 ± 4.1   & 68.4 ± 4.4  & 49.5 ± 4.6   & 75.3 ± 3.6 \\
LVMass$_i$ ($g/m^2$) & 62.8 ± 2.4 & 75.8 ± 6.1 & 74.3 ± 6.6   & 76.2 ± 11.8  & 87.3 ± 4.1   & 56.0 ± 1.9  \\
LVEF (\%)    & 64.1 ± 0.9  & 63.4 ± 2.1 & 59.1 ± 6.7   & 56.9 ± 3.5  & 58.3 ± 5.1   & 60.1 ± 0.9 \\ \hline
\end{tabular}%
}
\end{table}

Although all patients were included during the segmentation experiments, only those who underwent TTE and had a defined degree of LVDD were included in the subsequent quantitative and qualitative analysis. Controls and hypertensive patients with LVDD grades other than 0 or deemed indeterminate were excluded from the study, resulting in a final cohort comprising 68 patients. HCM patients were classified based on their LVDD gradation as follows: HCM with no LVDD (HCM), HCM with grade I LVDD (G1), and HCM with grade II LVDD (G2). Patients with grade II LVDD (G2) were further categorized according to the presence of systolic anterior motion (G2 - SAM), as the associated MV regurgitation induces sufficient alterations of LA hemodynamics to consider them as two separate groups. The clinical characteristics of the patients included in the subsequent analysis are summarized in Table \ref{table:Population}.

\subsection{Segmentation} \label{sec:Seg_results}

\begin{figure}[h!]
\centering
\includegraphics[width=0.8\textwidth]{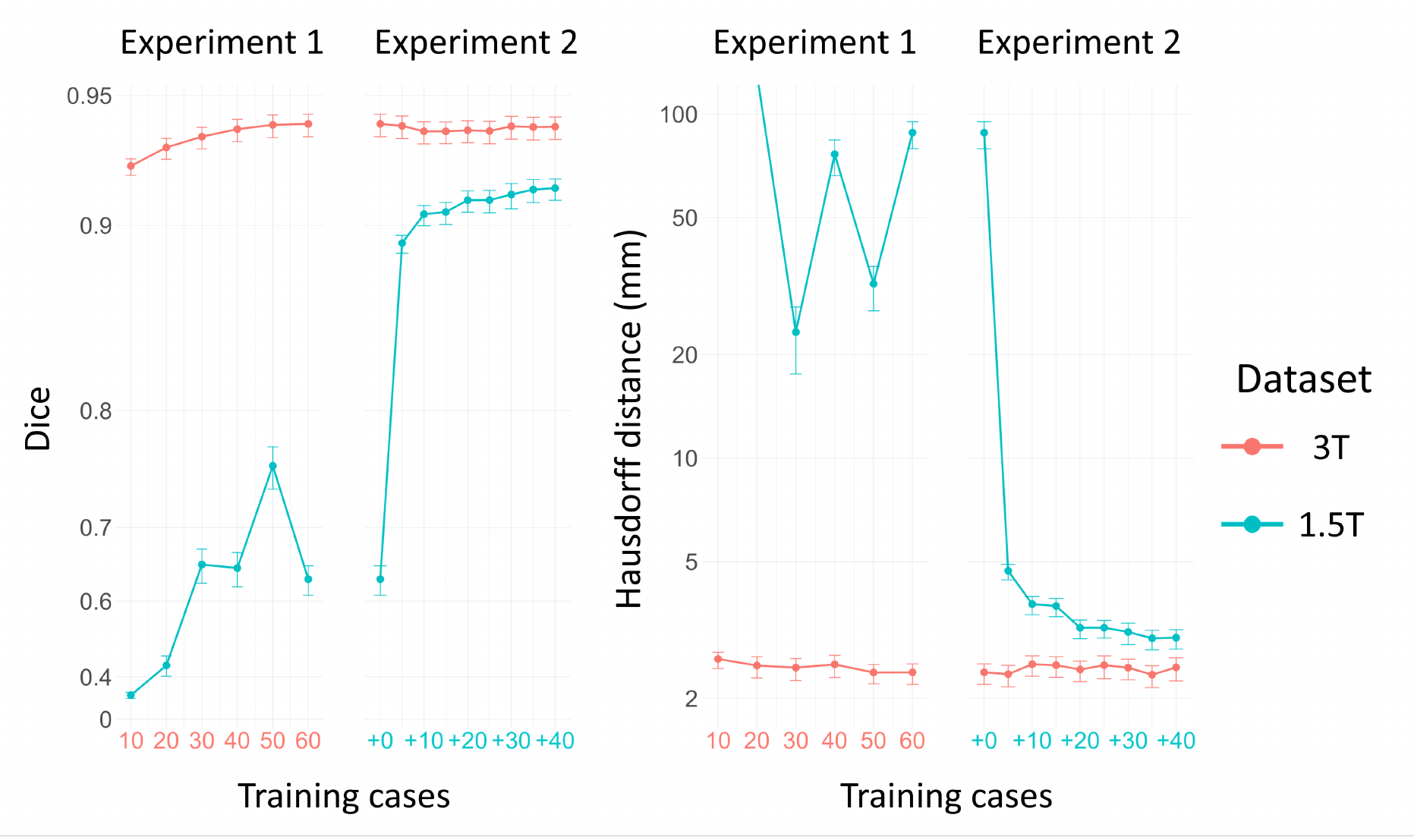}
\caption{Dice score and Hausdorff 95 distance (mm) for the segmentation experiments of the left atrium. For Experiment 1, the x-axis is the total number of training cases from dataset 3~T (in red), while in Experiment 2, it is the amount of 1.5~T cases (in blue) added on top of the complete 3~T training dataset.}
\label{fig:Segmentation}
\end{figure}

Figure \ref{fig:Segmentation} shows the segmentation accuracy for both experiments, while visualizations of the predicted segmentation contours for some representative examples are provided in Appendix \ref{segfigures}. In Experiment 1, an average Dice score of 0.924 and an HD95 distance of 2.61 mm were attained for 3~T cases from the outset, with only marginal improvements from additional training data. Conversely, accuracy was very poor in 1.5~T testing cases for both metrics, showing only a slight improvement in the Dice score with increasing training data. In Experiment 2, both metrics remained stable for the 3~T dataset, while the addition of the 5 1.5~T cases boosted the average Dice score from 0.635 to 0.892 and the HD95 distance from 88.5 mm to 4.70 mm in the low-resolution data. Although performance improved with additional 1.5~T training cases, the accuracy still did not match that of the 3~T dataset.

\subsection{4D Flow MRI-derived velocity spectrometry}

\begin{figure}[h]
\centering
\includegraphics[width=\textwidth]{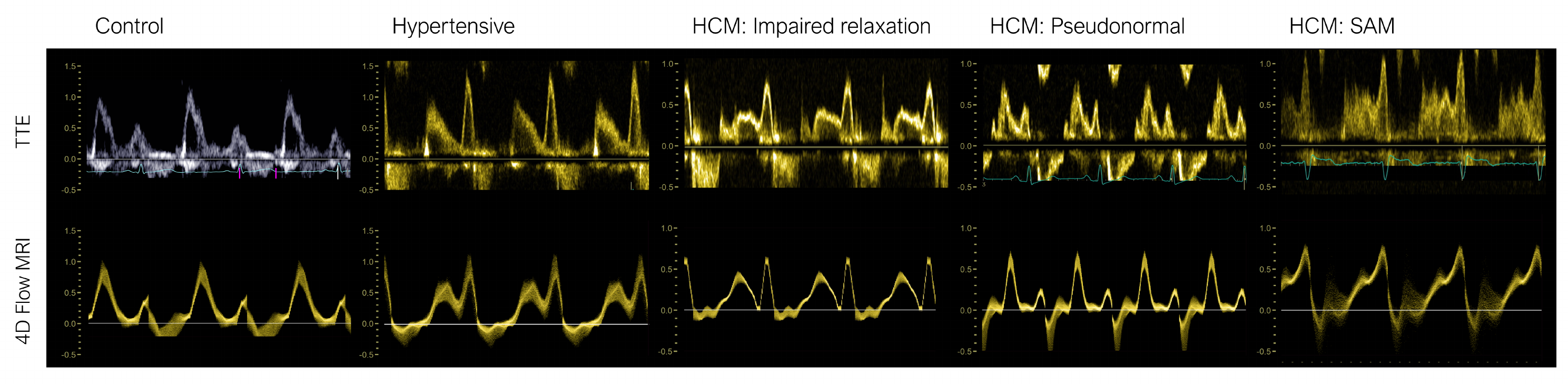}
\caption{Side-by-side comparison of conventional transthoracic echocardiography (TTE) pulsed-wave Doppler acquisitions and the 4D flow magnetic resonance imaging derived velocity spectrograms of the mitral valve. 4D Flow MRI returns a single cardiac cycle spectrogram that has been repeated for comparative purposes only. HCM: Hypertrophic cardiomyopathy; SAM: Systolic anterior motion.}
\label{fig:Spectro_MV}
\end{figure}

Ground truth MV spectrograms from TTE-based PW Doppler are shown alongside their 4D Flow MRI-derived counterparts in Figure \ref{fig:Spectro_MV}, displaying a range of distinct MV waveforms. The shape of the envelope was accurately preserved between the two modalities across different conditions. However, when compared to TTE, 4D Flow MRI underestimated E and A wave peaks by an average of -0.13 and -0.15 m/s, respectively.

\begin{figure}[h]
\centering
\includegraphics[width=\textwidth]{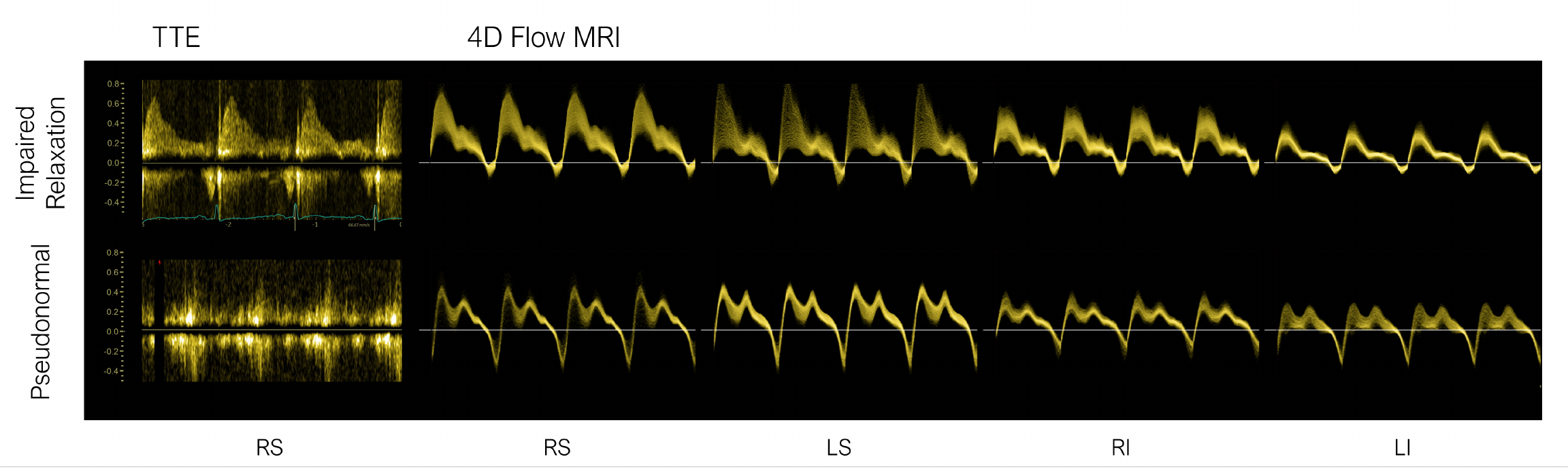}
\caption{Conventional transthoracic echocardiography (TTE) pulsed-wave Doppler acquisitions and 4D flow magnetic resonance imaging derived velocity spectrograms of the pulmonary veins (PV) from two hypertrophic cardiomyopathy patients. Both patients underwent TTE-based diastolic dysfunction grading \cite{Nagueh2016} and were found to be Grade I (impaired relaxation) and Grade II with a pseudonormal mitral flow pattern, shown in Figure \ref{fig:Spectro_MV}. Typically, only the right superior PV can be acquired in TTE, and the image can be of poor quality, as seen in the pseudonormal case. On the contrary, 4D Flow MRI allows easy measurement of the four PVs. 4D Flow MRI returns a single cardiac cycle spectrogram that has been repeated for comparative purposes only. RS: Right superior, LS: Left superior, RI: Right inferior, LI: Left inferior. }
\label{fig:Spectro_PV}
\end{figure}

Out of the 90 patients who underwent PW Doppler assessment, only 70 completed a PV acquisition, and just 50 had sufficient image quality for clinical analysis. An example of poor-quality PW Doppler is depicted in Figure \ref{fig:Spectro_PV}. Conversely, 4D Flow MRI enabled consistent measurement of all four PVs in all cases, though it maintained the tendency to underestimate peak values, both for the S wave (-0.12 m/s) and D wave (-0.13 m/s). This underestimation is particularly pronounced in the pulmonary reversal flow (-0.18 m/s), as illustrated in the impaired relaxation case in Figure \ref{fig:Spectro_PV}. 

\subsection{Left atrial volume}

\begin{figure}[h!]
\centering
\includegraphics[width=0.9\textwidth]{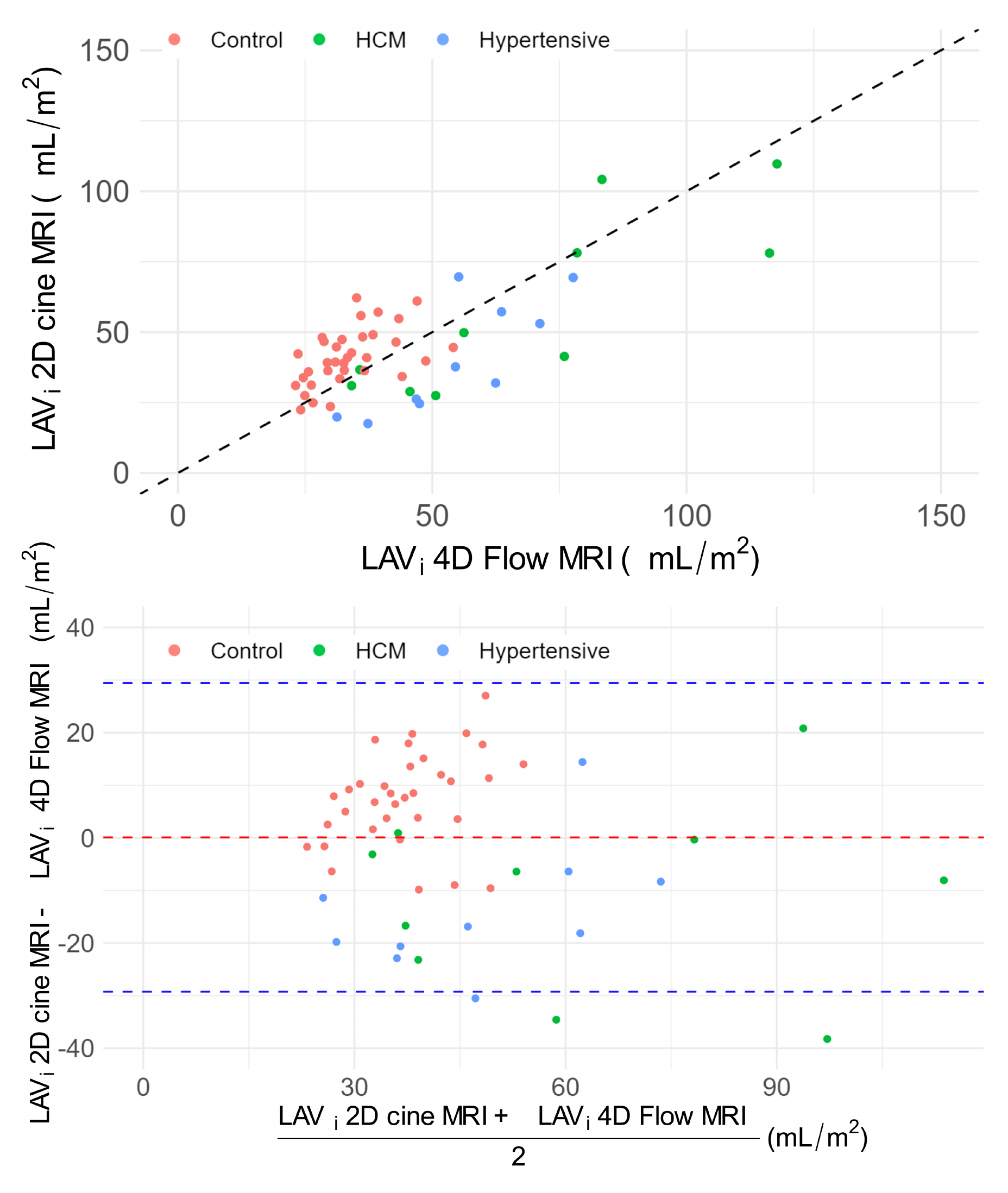}
\caption{Comparison of the maximal left atrium volume indexed by body surface area (LAV$_i$) derived from 2D cine and 4D flow magnetic resonance imaging. On top: scatter plot with the identity line for reference. Bottom: Bland-Altman plot with the bias as a dashed red line, while the 95\% confidence intervals are represented by dashed blue lines.}
\label{fig:Volume}
\end{figure}

The LAV$_{i}$ results are shown in Figure \ref{fig:Volume}. A strong correlation (Pearson correlation coefficient = 0.75, 95\% confidence interval 0.57-0.86) was observed between the 2D cine MRI-derived biplane method and the 3D PC-MRA-based LAV$_{i}$. The Bland-Altman analysis revealed a bias of 0.1 ml/m$^2$ and 95\% limits of agreement (LoA) of (-29.3, 29.4). However, when comparing both modalities for each cohort separately, a bias emerges, as evident in Figure \ref{fig:Volume}. A positive bias of 6.7 ml/m$^2$ was observed in controls (LoA: -12.9 to 26.3 ml/m$^2$) and a negative bias of -12.5 ml/m$^2$ for pathological cases (LoA: -41.8 to 16.8 ml/m$^2$).

\subsection{Flow Rate}

\begin{figure}[h!]
\centering
\includegraphics[width=0.8\textwidth]{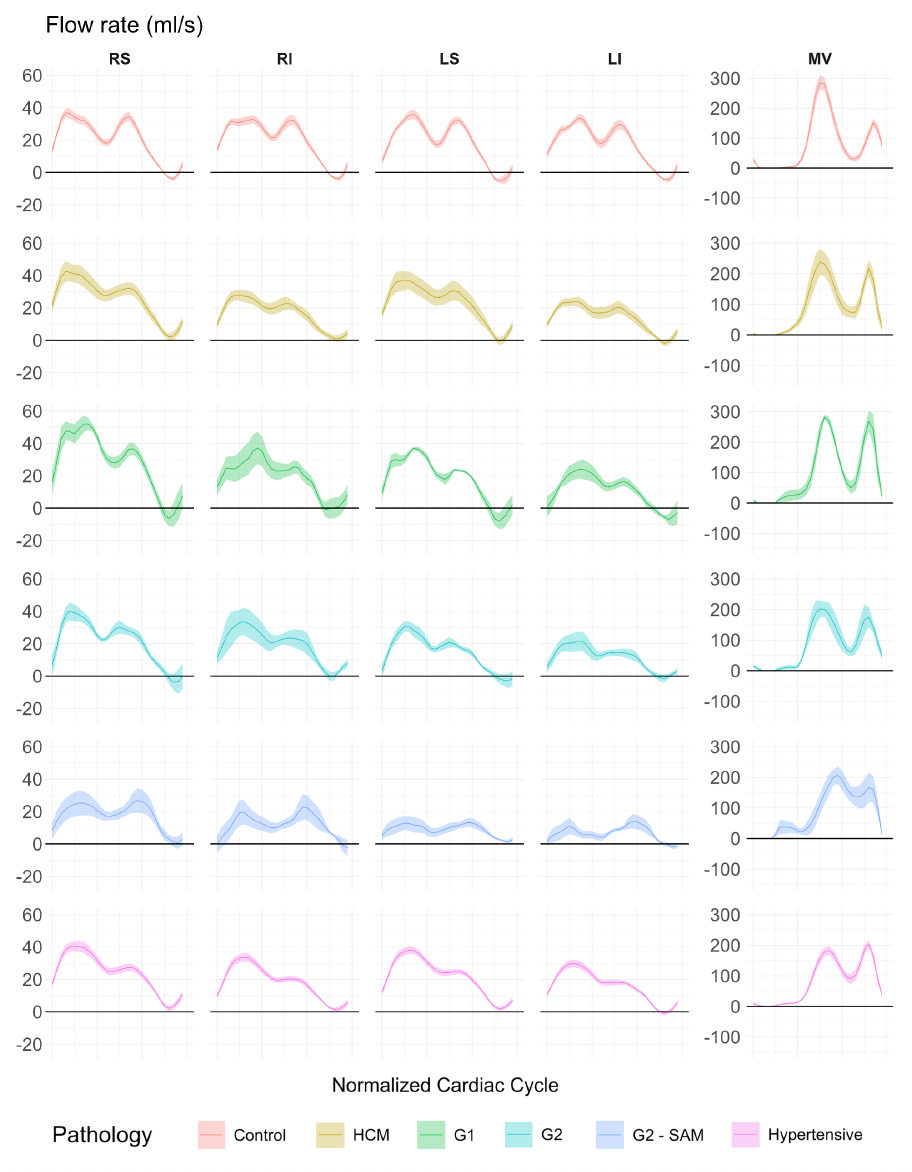}
\caption{Mean and standard error of the mean of the flow rate ($ml/s$) of the four pulmonary veins and the mitral valve. RS: Right superior; RI: Right inferior; LS: Left superior; LI: Left inferior; MV: Mitral valve; HCM: Hypertrophic cardiomyopathy with no left ventricular diastolic dysfunction (LVDD); G1: HCM with grade I LVDD; G2: HCM with grade II LVDD; G2 - SAM: HCM with grade II LVDD and systolic anterior motion of the mitral valve.}
\label{fig:FlowRate}
\end{figure}

Figure \ref{fig:FlowRate}, shows the average flow rate (ml/s) across the different pathology groups for the 4 PVs and the MV. Peak flow rate values are detailed in Table \ref{tab:FlowPeaks}. For the MV, the total volume expelled during passive and active filling was also computed.

The relationship between the four PVs varies by pathology. In healthy patients, flow peaks are fairly balanced among the veins. In pathological cases, however, the left inferior (LI) PV often shows the lowest flow rate. As shown in Table \ref{tab:FlowPeaks}, only the left PVs have significant differences between groups during the S wave. Although the post hoc analysis is not significant, G2 patients with SAM exhibit a notable decrease in flow in both left PVs. No significant differences were observed in the D wave, but left PV flow remains reduced in G2 - SAM patients.

The S/D ratio is relatively stable across the four PVs except in G2 - SAM patients. In the remaining pathological cases, the S/D ratio is significantly higher compared to healthy controls, with G1 and hypertensive patients exhibiting the highest ratios. The ANCOVA analysis is significant for all PVs except for the right superior. Regarding the peak PV reversal flow (Ar), all groups but G1 patients show smaller peak values than healthy controls. This difference is significant for all four PVs and is independent of age. Regarding the MV, the E/A ratios vary significantly whether computed based on peak flow rates or total volume, and are not linearly related.

\subsection{Energy quantification}

\begin{figure}[b!]
\centering
\includegraphics[width=0.9\textwidth]{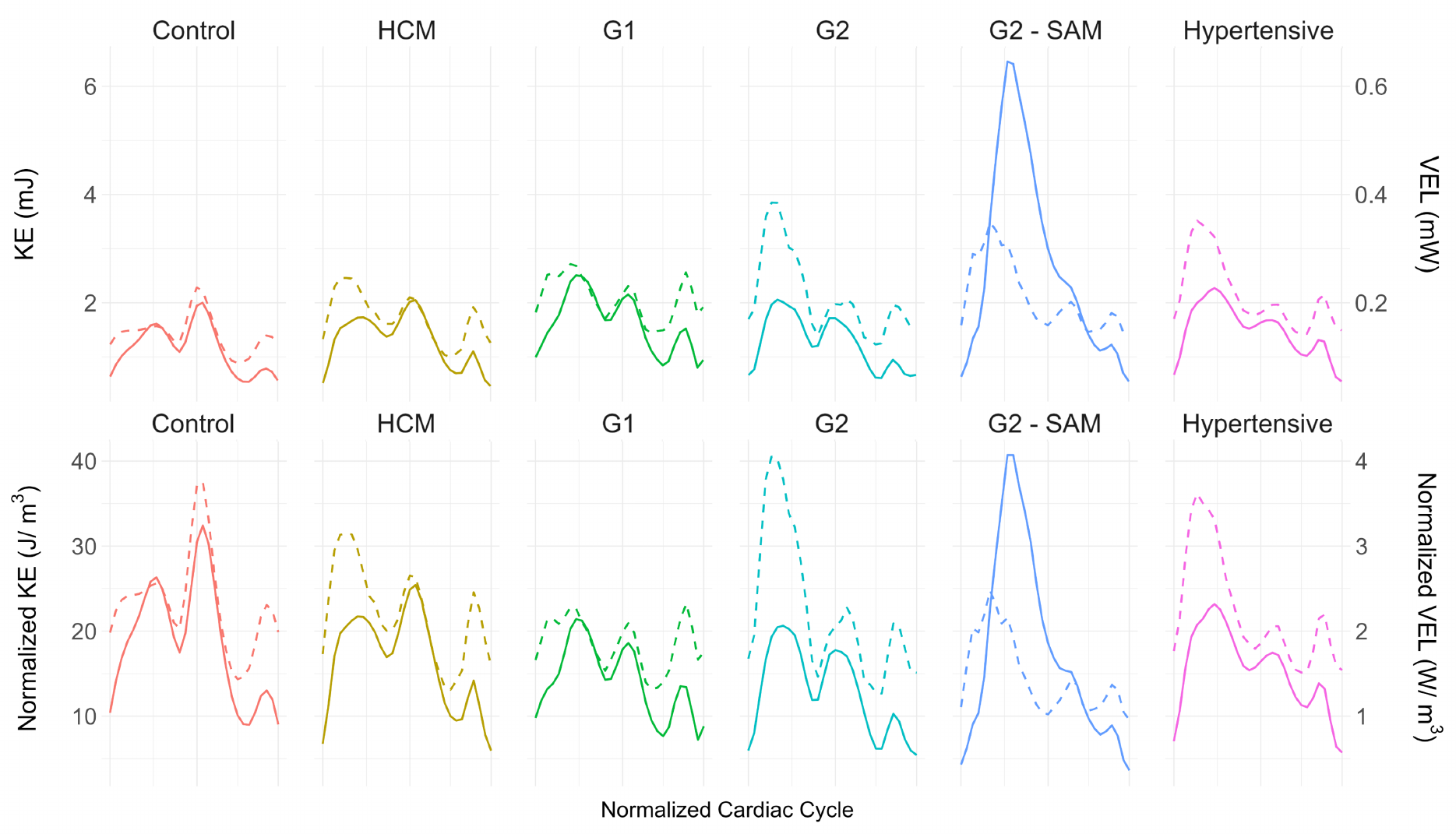}
\caption{Top: Mean kinetic energy (KE) as the solid line and mean viscous energy loss (VEL) as the dashed line in the left atrium (LA). Bottom: KE and VEL normalized by LA volume. HCM: Hypertrophic cardiomyopathy with no left ventricular diastolic dysfunction (LVDD); G1: HCM with grade I LVDD; G2: HCM with grade II LVDD; G2 - SAM: HCM with grade II LVDD and systolic anterior motion of the mitral valve.}
\label{fig:Energy}
\end{figure}

The mean kinetic energy (KE) and viscous energy loss (VEL) values are shown together in Figure \ref{fig:Energy}. After normalizing by LA volume, the average KE per unit volume becomes more equalized between groups, though its distribution over the cardiac cycle remains distinct. Both parameters show triphasic behavior, with an initial systolic peak related to pulmonary inflow during the S wave, followed by passive and active filling.

In healthy subjects and HCM patients, early diastolic KE predominates, even though there is a noticeable drop in the systolic and E wave peaks in the latter. In other pathological groups, the KE trend shifts toward a predominant systolic peak due to the noticeable decrease in early diastolic KE. According to Table \ref{tab:EnergyPeaks}, G2 patients with SAM exhibit significantly higher systolic peak values. Although substantial E-wave KE differences were observed across groups, these are age-dependent, according to the ANCOVA analysis. Lastly, the A wave KE peak is noticeably lower in both Grade II LVDD groups.
 
Regarding VEL, the early diastolic peak predominates once again for controls. However, contrary to KE, the systolic peak already predominates in HCM patients as well as in the remainder of pathological groups. HCM, G2, and hypertensive cohorts show elevated systolic VEL compared to other cohorts. Both systolic VEL and KE/VEL ratios show the most significant differences with a considerable effect size. Controls, G1 and especially G2 - SAM patients show the highest values.

\subsection{Vorticity} \label{sec: Vorticity}

\begin{figure}[H]
\centering
\includegraphics[width=0.9\textwidth]{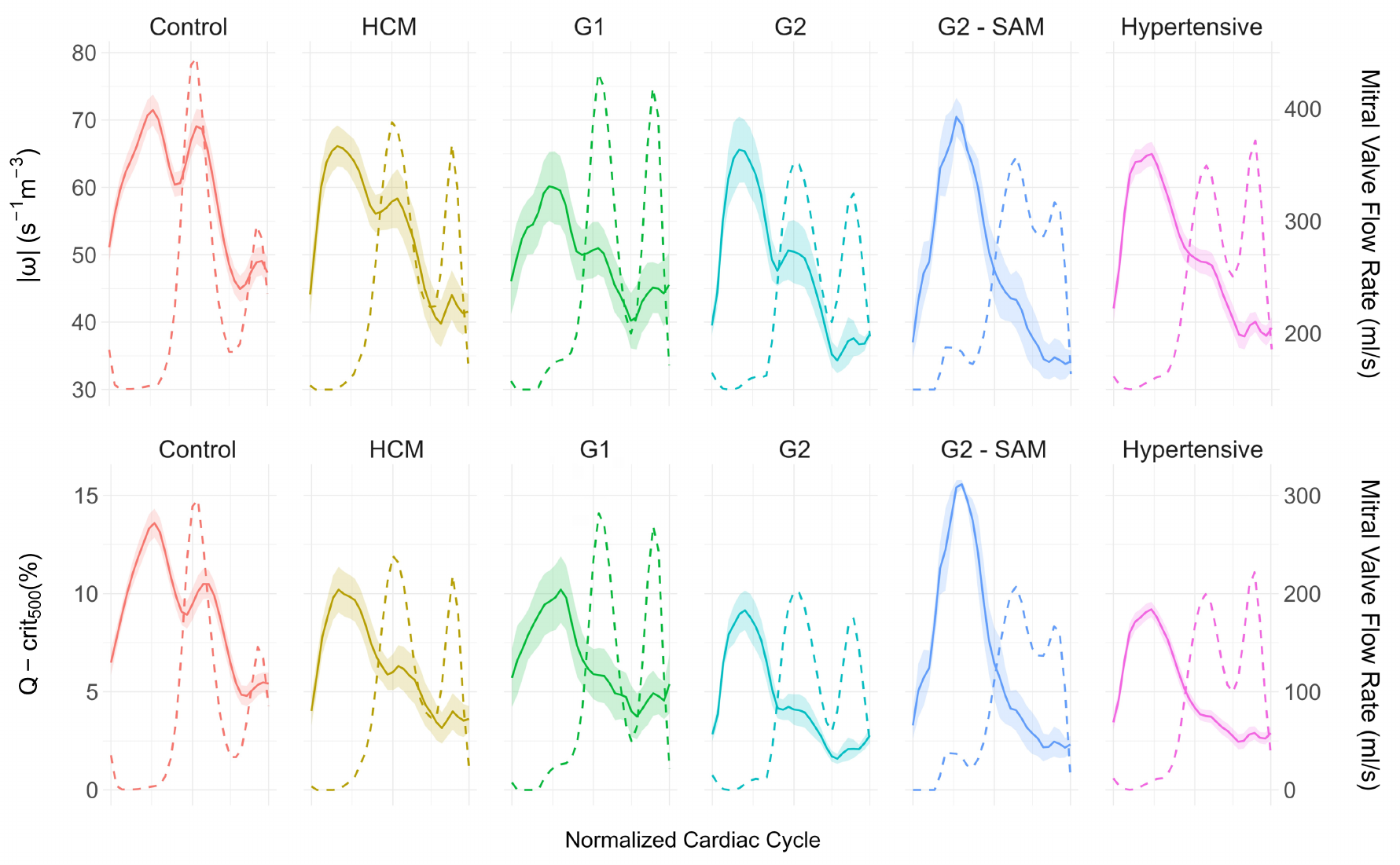}
\caption{Top: Mean and standard error of the mean of vorticity magnitude (solid line) with mitral valve flow rate (dashed line) for temporal reference. Bottom: Ratio of voxels in the left atrium with Q-criterion $>$ 500 s$^{-2}$ (solid line), associated with vortex core areas. HCM: Hypertrophic cardiomyopathy with no left ventricular diastolic dysfunction (LVDD); G1: HCM with grade I LVDD; G2: HCM with grade II LVDD; G2 - SAM: HCM with grade II LVDD and systolic anterior motion of the mitral valve.}
\label{fig:Vorticity}
\end{figure}

Figure \ref{fig:Vorticity} shows the temporal evolution of vorticity and the Q-criterion. Similar to the energy-based parameters, both vorticity measurements exhibit three peaks, though the A peak is almost negligible. Aside from systolic $\abs{\omega_{LA}}$, all other metrics exhibit strong statistical significance, independence from age, and substantial effect sizes (Table \ref{tab:VorticityPeaks}). G2 - SAM and control groups display the highest systolic vorticity, particularly in Q-crit${500}$. In controls, the early systolic and diastolic peaks are identical when measured with $\abs{\omega{LA}}$, but a noticeable gap appears when using Q-crit$_{500}$. In the remaining cohorts, the gap is reflected in $\abs{\omega{LA}}$, while it remains more prominent in Q-crit$_{500}$. The early diastolic peak is delayed relative to the E wave, peaking as the passive filling decelerates and it weakens progressively with LVDD severity. Vorticity is significantly lower during atrial contraction, with only G1 patients showing values comparable to controls.

\subsection{Relative pressure $\Delta P$}

\begin{figure}[h!]
\centering
\includegraphics[width=\textwidth]{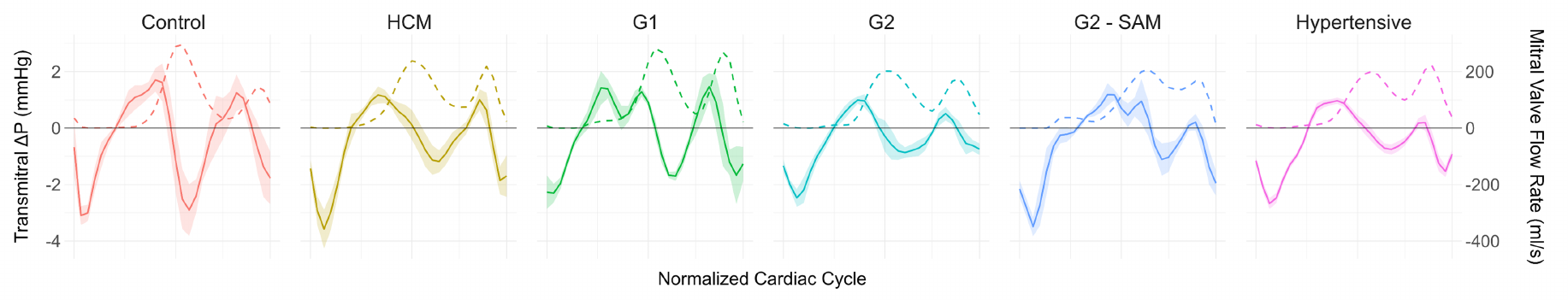}
\caption{Mean and standard error of the mean of the relative pressure (mmHg) between the left atrium and left ventricle, measured using the vWERP method (solid line), accompanied by the mitral valve flow rate (dashed line). HCM: Hypertrophic cardiomyopathy with no left ventricular diastolic dysfunction (LVDD); G1: HCM with grade I LVDD; G2: HCM with grade II LVDD; G2 - SAM: HCM with grade II LVDD and systolic anterior motion of the mitral valve.}
\label{fig:vWERP}
\end{figure}

As the pressure gradient between the LA and LV drives mitral flow, we integrated both measurements in Figure \ref{fig:vWERP}. In the Figure, positive values indicate higher pressure in the LA than in the LV, and vice versa. Two positive and two negative peaks are discernible during diastole, corresponding to the acceleration and deceleration of the E wave and A wave, respectively. Moreover, the saddle points align precisely with the points where the relative pressure crosses the zero line.

Both E peak acceleration ($\Delta$E$_{max}$) and deceleration ($\Delta$E$_{min}$), exhibit age-independent statistically significant differences with a substantial effect size (Table \ref{tab:PressurePeaks}). Controls have the highest pressure gradients by a significant margin, while differences among the pathological groups are more subtle. Although $\Delta$E${max}$ does not display a consistent pattern across pathological cohorts, G1 and G2 - SAM patients tend to have slightly higher values. Notably, in certain groups with LVDD, $\Delta$E$_{max}$ appears jagged, without a well-defined single maxima. The start of the E wave acceleration is also delayed in G2 - SAM patients. $\Delta$E$_{min}$ is also the highest in controls and progressively decreases with worsening LVDD. The A wave acceleration peak ($\Delta$A$_{max}$) mirrors vorticity, with higher values in controls and G1 patients compared to other cohorts. No significant differences are observed in the A wave deceleration peak ($\Delta$A$_{min}$).

\subsection{Visualization}

\begin{figure}[h!]
\centering
\includegraphics[width=0.97\textwidth]{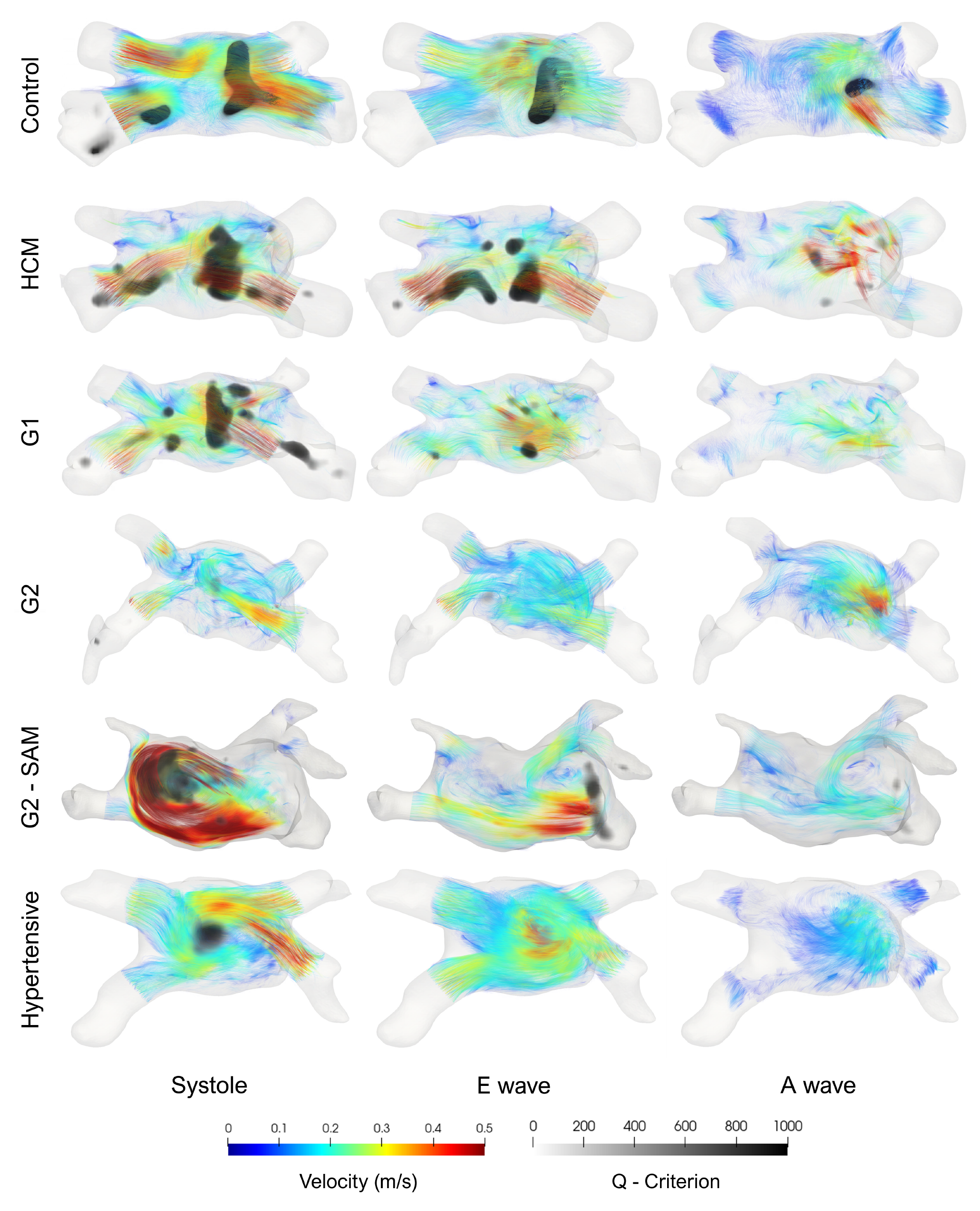}
\caption{An axial view of the central left atrial vortex across diverse cardiac disorders. The visualization combines the pathlines of particles emitted from the pulmonary veins, color-coded by velocity, and a volumetric rendering of the Q-criterion, indicating the core of the vortices. HCM: Hypertrophic cardiomyopathy with no left ventricular diastolic dysfunction (LVDD); G1: HCM with grade I LVDD; G2: HCM with grade II LVDD; G2 - SAM: HCM with grade II LVDD and systolic anterior motion of the mitral valve.}
\label{fig:Pathlines}
\end{figure}

Figure \ref{fig:Pathlines} depicts the central LA vortex of one case from each cohort at three points during the cardiac cycle corresponding to the Q-crit$_{500}$ peaks described in section \ref{sec: Vorticity}. The evolution of the vortex closely resembles the progression of the group averages shown in \ref{fig:Vorticity}. The vortex is strongest during systole across all patients but is notably weakened in G2 patients and particularly strong in G2 patients with SAM. After the E wave, the vortex remains robust in the control and is still noticeable in the HCM patient, but disappears in the remaining pathological cases. No coherent vortical structures are observed during atrial contraction.

\section{Discussion}

In this study, we have, to the best of our knowledge, developed the first framework for 4D Flow MRI analysis focused on the LA. We demonstrated the feasibility of automatic neural network-based segmentation of the LA over 4D Flow MRI-derived PC-MRA, allowing us to analyze a diverse cohort of patients from two different centers and vendors. This enabled a thorough quantitative and qualitative analysis of novel 4D flow MRI indices in the LA, marking the first time their values were reported in such a diverse cohort.

This pipeline aims to lay the groundwork for standardizing 4D Flow MRI analysis in the LA, allowing faster and more reproducible studies. Although some programming skills are still required, the goal is to reduce the skill barrier as much as possible, making advanced analysis more accessible to a wider audience. Additionally, automating the majority of time-consuming processes significantly reduces the analysis time per case from hours to minutes.

\subsection{Comparison with existing frameworks}

Currently, most 4D flow MRI analysis is conducted using specialized commercial software such as \href{https://www.mevis.fraunhofer.de/en/solutionpages/mevisflow-non-invasive-interactive-exploration-of-in-vivo-hemodynamics.html}{MEVISFlow} (MeVis Medical Solutions AG), \href{https://www.arterys.com/}{Arterys} (Tempus Labs), or \href{https://www.piemedicalimaging.com/product/mr-solutions/4d-flow}{CAAS MR 4D} (Pie Medical Imaging). However, free software alternatives are also available \cite{Khler2019, sotelo2019novel}. Although these tools often present a low-skill floor, they typically focus on a single cardiac structure and do not offer users the capability to introduce new quantitative or qualitative parameters. Considering that the analysis of 4D Flow MRI in the LA is still in its infancy,  this significantly curtails the exploration of innovative hemodynamic parameters and the ability to address specific clinical and physiological questions \cite{Heiberg2012}. 

Unfortunately, the adoption of fully open-source software in medical imaging lags behind other fields, such as deep learning. Beyond reproducibility and accessibility, open source is essential for a modality like 4D Flow MRI, which lacks standardization. DICOM files lack a common standard regarding image semantics, orientation, or ordering of the various components  \cite{Khler2019}, which often precludes the use of the analysis software if data deviates slightly from the expected format. To address this issue, our framework allows users to define custom loading functions before converting the data into a standardized RAS (Right, Anterior, and Superior) orientation. 

To our knowledge, FourFlow \cite{Heiberg2012} stands out as the sole fully open-source framework for 4D Flow MRI analysis. It offers a user-friendly interface based on the popular 3D viewer ParaView and supports customization via Python scripting, though this is limited to the tools provided by the Visualization Toolkit (VTK) within ParaView.  Additionally, some tasks such as data loading and segmentation, require the use of Segment \cite{Heiberg2010}, which primarily relies on MATLAB, a proprietary programming language. 

Conversely, all data processing, automatic segmentation, and quantitative analysis in our framework are conducted purely in Python, which enables leveraging the ecosystem of the most widely used open-source data analysis programming language. Manual segmentation and 3D visualization tasks are delegated to open-source programs like Slicer 3D and ParaView, but we provide Python scripts to automate the most repetitive and time-consuming tasks. As noted in Section \ref{sec:postpro}, the only user input required is the placement of sample volume spheres within ParaView. While the detection of PV and MV can be automated, the heterogeneity of the LA would make it highly prone to error. 

Consequently, as long as segmentation is automated through neural networks, the analysis of a new case through our framework can take as little as 10 minutes. However, framework customizability comes at the cost of increasing the learning curve. At the moment, our pipeline requires a minimum proficiency in programming and 3D software, which may render it less appealing to people without a technical background.

\subsection{Segmentation}

The results presented in Section \ref{sec:Seg_results} evidence that a small dataset with high-quality annotations may suffice to automate 4D Flow MRI segmentation. Prior studies have similarly shown that neural networks can achieve accurate segmentation with limited data \cite{BarreraNaranjo2023}, challenging the presumed advantages of alternative approaches such as atlas-based segmentation \cite{Bustamante2018}.

Experiment 2 underscores the network's adaptability to generalize across diverse datasets. The 1.5~T data originates from a different vendor, center, and protocol, and has undergone upsampling-denoising, resulting in a radically different distribution from the 3~T dataset. This is evident in the lower accuracy observed for the 1.5~T data in Experiment 1, as shown in Figure \ref{fig:Segmentation}. Notably, adding just 5 cases from the 1.5~T dataset significantly improved results without compromising the accuracy of 3~T data. In the context of 4D Flow MRI, where data is often limited, this approach could substantially expedite the compilation of multicenter datasets for more comprehensive clinical studies.

\subsection{Left atrial volume}

A strong correlation was found between the 2D cine and 4D Flow MRI-derived LAV$_i$ with little to no bias and limits of agreement that were consistent with previous studies comparing the biplane method with 3D acquisitions \cite{Maroun2023}. However, a notable shift was observed in the behavior of the bias between control subjects and pathological patients. Several factors could contribute to this disparity. The 2-chamber and 4-chamber 2D cine images are not always perfectly aligned with the LA axes, unlike 4D Flow MRI, which encompasses all cardiac chambers and thus is not susceptible to such errors. Additionally, the geometrical assumptions underlying the biplane method fail to consider the changes in LA shape caused by remodeling \cite{Blume2011}. Similar underestimation of pathological cases by the biplane method has been reported for patients with atrial fibrillation \cite{Maroun2023}.

Although not its primary purpose, we have shown that 4D Flow MRI can reasonably estimate LA volume. Further studies are needed to validate its accuracy against reference 3D sequences like CT. While our focus was on LAV$_{i}$ given the use of a static 3D PC-MRA, time-resolved segmentation could enable analysis of LA function by tracking volume changes throughout the cardiac cycle. 4D Flow MRI is unlikely to become a reference method for volume quantification due to spatiotemporal resolution limitations, but no other imaging modality can simultaneously assess LA size, function, and hemodynamics. Lastly, despite the recognized drawbacks of 2D methods, the demand for specialized expertise and software has hindered wider clinical adoption of 3D modalities. We postulate frameworks like ours can foster broader acceptance.

\subsection{4D Flow MRI-derived velocity spectrometry}

Figures \ref{fig:Spectro_MV} and \ref{fig:Spectro_PV} illustrate the feasibility of generating velocity spectrograms from 4D Flow MRI data, allowing the assessment of conventional flow parameters typically measured by TTE. Despite differences in spatio-temporal resolution and acquisition modality, velocity spectrograms from PW Doppler TTE and 4D Flow MRI exhibit similar envelopes across various conditions in both MV and PVs.

While TTE is constrained by geometry, depth, and operator expertise, 4D Flow MRI provides retrospective measurements at any arbitrary location within the acquisition volume. This is particularly valuable for assessing the PVs, which are often difficult to image with TTE due to their far-field location \cite{Huang2008}. Only half of the patients in our study had a satisfactory PV PW Doppler spectrogram from TTE. Moreover, TTE is typically limited to measuring only the right superior PV, whereas 4D Flow MRI allows for simultaneous assessment of all four PVs. Figure \ref{fig:Spectro_PV} demonstrates that each of these veins has unique dynamics, potentially providing new insights into LA hemodynamics.

Thus, 4D Flow MRI can offer a more reproducible and operator-independent PV analysis, which is particularly valuable in the non-invasive evaluation of LVDD. For instance, the Ar-A duration, which compares the duration of pulmonary venous and mitral valve A wave velocities, is an ejection fraction and age-independent criterion associated with elevated left ventricular end-diastolic pressures, commonly used in the grading of LVDD in HCM patients \cite{Nagueh2016}. In this regard, Figure \ref{fig:Spectro_MV} and Figure \ref{fig:Spectro_PV} highlight the added value that 4D Flow MRI can provide. In pseudonormal patients (LVDD Grade II), the increase in atrial pressure typically normalizes the mitral flow pattern despite the presence of impaired relaxation. As seen in Figure \ref{fig:Spectro_MV}, focusing solely on MV flow can make it difficult to distinguish a healthy subject from a pseudonormal patient. The Ar-A duration is one of the key measurements that can help identify pathological cases, however, as shown in Figure \ref{fig:Spectro_PV}, the poor quality of the TTE PV acquisition prevents its correct measurement.

This underscores the potential of 4D Flow MRI, not as a replacement, but as a complementary tool for conventional flow analysis. Velocity spectrograms derived from 4D Flow MRI could facilitate obtaining the standardized TTE measurements necessary to complete guideline-based cardiac assessment in patients with poor acoustic windows, such as those who are obese or have chronic lung disease. Furthermore, linking 4D Flow MRI to well-established TTE functional assessment measures could be a significant step toward validating 4D Flow MRI for broader clinical use, fostering greater confidence and acceptance within the clinical community.

Finally, it is worth noting that 4D Flow MRI-derived spectrograms consistently exhibited a noticeable underestimation of peak values compared to TTE. This discrepancy can be attributed to the nature of 4D Flow MRI acquisition, where velocity fields are averaged over multiple cardiac cycles, often with varying RR intervals \cite{Dyverfeldt2015}. This can lead to the over-smoothing of transient, short-lived instantaneous patterns over a prolonged data acquisition process. Consequently, this effect is most noticeable in the pulmonary reversal or Ar peak, as shown in the impaired relaxation case in Figure \ref{fig:Spectro_PV}. If this underestimation of velocities is found to be consistent, correction factors could be developed, but they would require independent validation for each center, vendor, and protocol.

\subsection{Flow rate}

Unlike conventional TTE, which is primarily used to measure peak velocities, 4D Flow MRI can also quantify flow rates and volumes without making assumptions about the flow profile. By integrating the total volume of fluid passing through a vessel cross-section, flow rate measurements are more robust and less susceptible to data noise. Notably, flow rate peaks have been shown to outperform TTE velocity peaks in the computation of E/A ratios for the evaluation of LVDD \cite{Alattar2022}. A key limitation of peak velocities is that they don't account for changes in vessel cross-sectional area. For instance, PV distensibility can be affected by pulmonary circulation alterations or postural changes \cite{Wieslander2019}, which impact the maximum observed velocity even though the flow rate remains constant.

The statistical analysis in Table \ref{tab:FlowPeaks} highlights significant differences in the peak E/A ratios derived from flow rates and total volumes. Flow rates, while more robust than velocity values, represent only a single snapshot of the cardiac cycle. In contrast, volumes capture the entire filling phase, offering a more comprehensive assessment of LVDD by reflecting the real ratio of passive to active filling phases.

We hypothesize that PV flow rates could also provide more reliable measurements of parameters such as the S and D peaks, the S/D ratio, and pulmonary reversal flows compared to conventional peak velocity measurements. The absence of cases with an S/D ratio lower than one likely reflects the limited number of advanced LVDD cases in the dataset. Notably, the RS vein shows the least variation between different pathologies, while the other three PVs, seldom measured with TTE, exhibit the most statistically significant differences and largest effect sizes. This raises the question of whether we are overlooking relevant diagnostic information by focusing solely on the RS PV in clinical practice \cite{Blume2011}. Finally, patients with SAM exhibit a marked difference between the left and right PVs during the S and D wave, supporting the hypothesis that the mitral regurgitation jet, which is oriented postolaterally, obstructs the left pulmonary inflow.

\subsection{Energy quantification}

The temporal evolution of KE in the LA has been previously reported in three studies \cite{Arvidsson2013,Gaeta2018,Gupta2021}, involving a limited number of healthy subjects and HCM patients, showing results similar to those in Figure \ref{fig:Energy}. The predominance of the early diastolic peak reflects a compliant LV in healthy subjects and, to a lesser extent, in HCM patients without LVDD. However, in the remaining four groups, the KE curve shifts toward a predominant systolic peak. This shift is particularly pronounced in SAM patients due to mitral regurgitation, which results in a characteristic rotational flow pattern shown in Figure \ref{fig:Pathlines}. In Grade I and II LVDD and hypertensive patients, the predominance of systolic KE appears to be related to an increase in the S/D ratio.

The KE/VEL ratio can serve as a measure of flow efficiency, where a higher ratio indicates that more energy is utilized to transport the fluid, with less energy lost to friction. The notable increase in the KE/VEL ratio during systole in both controls and G2-SAM patients may be linked to increased vorticity, as shown in Table  \ref{tab:VorticityPeaks}, since rotational flow better preserves KE \cite{Arvidsson2013}. However, it is unclear why systolic flow efficiency appears to be higher in G1 than in HCM or G2 patients. No significant differences were observed in the KE/VEL ratio between groups during diastole, but the notable overall decrease observed between the E wave and A wave suggests that active filling is less efficient in conserving momentum, beyond the energy needed for atrial contraction.

\subsection{Vorticity}

Knowledge of rotational flow in the LA dates back to 2001 \cite{Fyrenius2001}. Vortices are thought to preserve momentum \cite{Arvidsson2013} and may help prevent thrombus formation \cite{Garcia2019}. Importantly, vortex structures are sensitive markers of heart disease, making the study of vorticity highly compelling for detecting early changes in LA function \cite{Suwa2014,Garcia2019,Sekine2022}.

The optimal approach to detect vortical structures remains contentious in fluid dynamics \cite{Gnther2018}, including in 4D Flow MRI. A key consideration is whether to normalize vorticity metrics by total volume, as simply summing vorticity across all voxels biases results in patients with dilated LAs. For instance, \cite{Garcia2019} reported a positive correlation between vortex size and both age and LA remodeling, even though LA function typically declines with age. Similarly, Spartera et al. \cite{Spartera2021} found that AF patients had larger vortex sizes compared to those in sinus rhythm, despite the weakened S wave and absence of diastasis in fibrillation patients, the only phases in the cardiac cycle where the atrial vortex is visible. In addition, although often not explicitly stated \cite{Schfer2017,Spartera_2023}, it is also common to measure vorticity based on magnitude alone, which can be misleading. Increased velocity along a non-straight flow path will invariably elevate the magnitude of vorticity, introducing a spurious term unrelated to the presence of a vortex. 

Parameters such as the Q-criterion and $\lambda_2$ were explicitly designed to address these limitations and distinguish rotational from strain-dominated regions. More specifically, the Q-criterion identifies vorticity-dominant areas and enables the visualization of different scale structures by adjusting the threshold value, where a higher threshold emphasizes large-scale vortex cores. The threshold value of 500 was empirically chosen based on the visualizations in Figure \ref{fig:Pathlines}, but the results were consistent for all values above 100.

We attribute the discrepancies observed between Q-Crit$_{500}$ and $\abs{\omega_{LA}}$, particularly during the early filling peak, to the term specifically linked with the increase in velocity. Supporting this, we found that peak $\abs{\omega_{LA}}$ occurs, on average, 15 ms earlier (cardiac cycle normalized to 60 beats per minute) and is closer in time to the E flow rate peak. Visualizations in Figure \ref{fig:Pathlines} also align more closely with the Q-Crit$_{500}$ curves in Figure \ref{fig:Spectrogram}. For instance, although G2 patients with SAM exhibit the most visually intense rotational flow during systole due to the mitral regurgitation jet, according to $abs{omega_{LA}}$ values in Table \ref{tab:VorticityPeaks}, their vorticity is not higher than that of the controls. Similarly, $\abs{\omega_{LA}}$ does not capture the generalized weakening of the vortices from systole to early diastole as effectively as Q-Crit$_{500}$.

The primary hypothesis for vortex generation in the LA is the collision of PV influxes. Our findings reinforce this notion, as the two main vorticity maxima coincide precisely with the peaks of pulmonary flow rate. Conversely, we observe a significant reduction in vorticity during atrial contraction, which actively opposes pulmonary inflow.

By utilizing a specialized vortex detection metric normalized by the LA volume, we effectively quantify the temporal evolution of LA vorticity, correlating well with previous qualitative analyses \cite{Suwa2014}. Most importantly, vorticity measurements, and particularly the Q-Crit$_{500}$, exhibit statistically significant differences between groups, irrespective of age. The early diastolic vorticity peak is of particular interest for the non-invasive assessment of LVDD, as it appears to gradually deteriorate with worsening diastolic dysfunction.

\subsection{Relative pressure $\Delta P$}

While transmitral relative pressure had been previously estimated using vWERP in 4D Flow MRI data \cite{Marlevi2021}, the study was limited to a small cohort with a single LVDD patient \cite{Marlevi2021}. Our study represents the first comprehensive evaluation of vWERP in the left heart, encompassing a larger and more diverse patient population, including various stages of LVDD. As expected, healthy subjects show the highest $\Delta$E${max}$ and $\Delta$E${min}$, reflecting a compliant LV that relaxes quickly and efficiently. The elevated $\Delta$E$_{max}$ in G2-SAM patients may be linked to mitral regurgitation, which distends the LA and delays passive filling initiation, resulting in a steeper initial pressure gradient.

In our limited cohort, $\Delta$E$_{min}$ seems to better reflect LVDD progression than $\Delta$E$_{max}$. $\Delta$E$_{min}$ can be considered analogous to the mitral flow deceleration time (DT), a Doppler TTE-derived parameter often included in LVDD assessment\cite{Nagueh2016}. Unlike DT, which is not recommended in patients with E and A wave fusion (e.g., hypertensives and G2-SAM patients in our study), $\Delta$E${min}$ can easily identify the end of the E wave by pinpointing the moment when the pressure gradient crosses the zero line, as shown in Figure \ref{fig:vWERP}.

Lastly, $\Delta$A$_{max}$ shows the expected progression across different LVDD stages in HCM patients. A significant increase is seen from HCM to G1, indicating an initial boost in atrial contractility to compensate for the impaired relaxation. Afterward, $\Delta$A$_{max}$ appears to progressively decrease with the deterioration of atrial function in G2 and G2 - SAM patients. 

Although methods like vWERP or hemodynamic forces \cite{Arvidsson2017} offer promising alternatives to invasive catheter-based measurements, they remain proxies for absolute pressures. The nonlinear changes in E and A waves during LVDD progression can create ambiguity, especially without knowledge of absolute LA or LV filling pressures. This can make it challenging to differentiate between different grades of LVDD, or between a healthy individual and a patient with a pseudonormal filling pattern. Therefore, the true clinical impact and predictive power of 4D Flow MRI-derived atrioventricular pressure gradients need further evaluation in dedicated studies.

\subsection{Limitations}

A major limitation of this study is the lack of temporally resolved segmentation. While dynamic LA segmentation using time-resolved PC-MRA from \cite{Bustamante2017} is likely feasible during cardiac intervals with high velocities, the low signal-to-noise ratio in the LA during other intervals can make single-step segmentation challenging and non-reproducible. Additionally, this approach would significantly increase the annotator's workload by 20 to 30 times. Consequently, previous attempts have relied on manually segmented end-diastolic and end-systolic frames, combined with non-rigid registration to estimate the remaining frames, which raises concerns about the accuracy of intermediate timesteps \cite{Bustamante2018}. Incorporating 4D convolutions to make the segmentation networks aware of temporal consistency, along with the use of sparse loss functions, can reduce the manual segmentation burden to just 20\% of the total dataset \cite{myronenko20204d}, while making the prediction of intermediate timesteps more reliable. 

Registration of 4D Flow MRI to sequences with better myocardium-to-blood contrast sequences, such as 3D bSSFP MRI, is another viable alternative \cite{Corrado2022}. However, given that bSSFP is not always available and considering the overarching goal of integrating multicenter data, we chose to focus on the magnitude and phase images, which are an intrinsic part of 4D Flow MRI acquisitions. Finally, time-resolved segmentation would also require additional preprocessing, such as the need to track the mitral valve and the four PVs for accurate flow rate measurement \cite{Juffermans2021}.

Another disadvantage of using static segmentation is the possibility of including static tissue, which would violate the continuity assumptions of the Navier-Stokes equations and increase the vWERP estimation error. However, vWERP has demonstrated reliable performance in assessing intracardiac pressure gradients within dynamic flow domains, even with a static mask \cite{Marlevi2021}, provided that the mask approximates an intersection segmentation (i.e. a segmentation that remains within the flow domain at all times). The time-averaged PC-MRA is biased towards regions of signal accumulation where there is a constant flow and vWERP works reliably. Although some static tissue may be included at the edges, previous studies have shown that minor segmentation dilations have negligible effects on pressure gradient estimation \cite{Marlevi2021,marlevi2021false}.

Regarding the denoising and upsampling of 4D Flow MRI data, 4DFlowNet was originally trained with synthetic CFD data from the aorta and tested in single-center, single magnetic field strength in vivo data. Ideally, 4DFlowNet should be validated using paired low- and high-resolution 4D Flow MRI acquisitions, but obtaining such data is extremely challenging. Alternatively, retraining with domain-specific CFD simulations has been shown to reduce the prediction error \cite{ferdian2023cerebrovascular}. However, despite advances in wall motion boundary conditions based on dynamic CT acquisitions \cite{albors2022sensitivity}, CFD simulations of the LA are still not realistic enough \cite{morales20214d} to provide a significant advantage over the original network. 

On the other hand, caution should be exercised in interpreting the significance of the statistical analysis due to the small cohorts, especially concerning the different grades of LVDD. The main objective of this preliminary study is to illustrate the potential of the framework and report benchmark values of novel 4D Flow MRI indices in the LA, laying the groundwork for future research. By making the entire process publicly available, we hope to facilitate the collection of larger 4D flow MRI datasets in the LA, which could help corroborate or challenge our findings. It is worth noting that the most statistically significant differences were observed in parameters exclusive to 4D flow MRI. Current non-invasive guidelines for the assessment of LVDD have been reported to have a sensitivity as low as 35\% \cite{vandeBovenkamp2021}, underscoring the need to identify new parameters for assessing diastolic relaxation abnormalities independently of confounding factors such as age \cite{MacNamara2021}.

However, given the heterogeneity in the etiology of disorders like LVDD, it is unlikely that a single parameter will provide a definitive diagnosis. In addition, much of the quantitative analysis in this study, as well as most clinical guidelines in the diagnosis of cardiac disorders, rely on measuring discrete peak values at specific points in the cardiac cycle, of inherently time-resolved signals. Moreover, modalities like 4D flow MRI generate vast amounts of data that exceed human interpretation capabilities. The incorporation of unsupervised machine learning methods could better integrate the wealth of data generated by the framework, as they can extract underlying patterns from time-resolved, high-dimensional data to create interpretable and clinically relevant phenogroups \cite{Loncaric2021}. 

\section{Conclusions}

In the present study, we present the first fully open-source computational pipeline for the comprehensive analysis of multiple center 4D Flow MRI data in the LA. LA segmentation was fully automated, leveraging well-established neural networks, and generalizing to data from two distinct centers, even with a limited number of training cases. We generated 4D Flow MRI-derived velocity spectrograms that address the geometry and depth limitations of conventional TTE PW Doppler. Furthermore, a comprehensive analysis of novel 4D Flow MRI indices in the LA was conducted in a diverse cohort of distinct disorders, revealing several potentially relevant pathophysiological trends. By making the entire process open source, we hope to encourage further research into LA hemodynamics and contribute to standardizing and accelerating its analysis.

\section*{Funding}

This study was supported by the Agency for Management of University and Research Grants of the Generalitat de Catalunya under the Grants for the Contracting of New Research Staff Programme - FI (2020 FI\_B 00608) and the Spanish Ministry of Economy and Competitiveness under the Programme for the Formation of Doctors (PRE2018-084062), the Maria de Maeztu Units of Excellence Programme (MDM-2015-0502) and the Retos Investigación project (RTI2018-101193-B-I00). Additionally, this work was supported by the H2020 EU SimCardioTest project (Digital Transformation in Health and Care SC1-DTH-06-2020; grant agreement No. 101016496). This research was also supported by grants from NVIDIA and utilized NVIDIA RTX A6000. Data collection and analysis at the University of Auckland is supported by the Health Research Council of New Zealand (program grant 17/608). Funded also by Instituto de Salud Carlos III, Gobierno de España grant FIS PI21/00905 and Hospital Clinic, Premi Emili Letang-Josep Font (MR). D.M. acknowledges funding from the European Union (ERC, MultiPRESS, 101075494). Views and opinions expressed are those of the authors and do not reflect those of the European Union or the European Research Council Executive Agency. 

\section*{Data Availability Statement}

The code for the computational pipeline is publicly available and can be accessed at the following repository: \href{https://github.com/Xtaltec/LA-4D-Flow-MRI}{github.com/Xtaltec/LA-4D-Flow-MRI}.



\clearpage

\bibliographystyle{unsrtnat}
\bibliography{references}  

\begin{thebibliography}{62}
\providecommand{\natexlab}[1]{#1}
\providecommand{\url}[1]{\texttt{#1}}
\expandafter\ifx\csname urlstyle\endcsname\relax
  \providecommand{\doi}[1]{doi: #1}\else
  \providecommand{\doi}{doi: \begingroup \urlstyle{rm}\Url}\fi

\bibitem[Fyrenius(2001)]{Fyrenius2001}
A~Fyrenius.
\newblock Three dimensional flow in the human left atrium.
\newblock \emph{Heart}, 86\penalty0 (4):\penalty0 448--455, October 2001.
\newblock \doi{10.1136/heart.86.4.448}.
\newblock URL \url{https://doi.org/10.1136/heart.86.4.448}.

\bibitem[Inciardi and Rossi(2019)]{inciardi2019left}
Riccardo~M Inciardi and Andrea Rossi.
\newblock Left atrium: a forgotten biomarker and a potential target in cardiovascular medicine.
\newblock \emph{Journal of Cardiovascular Medicine}, 20\penalty0 (12):\penalty0 797--808, 2019.

\bibitem[Fang et~al.(2022)Fang, Li, Wang, Wang, Allen, Ching, Zhong, and Li]{Fang2022}
Runxin Fang, Yang Li, Jun Wang, Zidun Wang, John Allen, Chi~Keong Ching, Liang Zhong, and Zhiyong Li.
\newblock Stroke risk evaluation for patients with atrial fibrillation: Insights from left atrial appendage.
\newblock \emph{Frontiers in Cardiovascular Medicine}, 9, August 2022.
\newblock \doi{10.3389/fcvm.2022.968630}.
\newblock URL \url{https://doi.org/10.3389/fcvm.2022.968630}.

\bibitem[MacNamara and Sarma(2021)]{MacNamara2021}
James~P. MacNamara and Satyam Sarma.
\newblock Faltering under pressure: Limitations to noninvasive diastolic function assessments.
\newblock \emph{Journal of the American Heart Association}, 10\penalty0 (18), September 2021.
\newblock \doi{10.1161/jaha.121.023189}.
\newblock URL \url{https://doi.org/10.1161/jaha.121.023189}.

\bibitem[Ashkir et~al.(2022)Ashkir, Myerson, Neubauer, Carlh\"{a}ll, Ebbers, and Raman]{Ashkir2022}
Zakariye Ashkir, Saul Myerson, Stefan Neubauer, Carl-Johan Carlh\"{a}ll, Tino Ebbers, and Betty Raman.
\newblock Four-dimensional flow cardiac magnetic resonance assessment of left ventricular diastolic function.
\newblock \emph{Frontiers in Cardiovascular Medicine}, 9, July 2022.
\newblock \doi{10.3389/fcvm.2022.866131}.
\newblock URL \url{https://doi.org/10.3389/fcvm.2022.866131}.

\bibitem[Marino et~al.(2019)Marino, Degiovanni, and Zanaboni]{Marino2019}
Paolo~N Marino, Anna Degiovanni, and Jacopo Zanaboni.
\newblock Complex interaction between the atrium and the ventricular filling process: the role of conduit.
\newblock \emph{Open Heart}, 6\penalty0 (2):\penalty0 e001042, October 2019.
\newblock \doi{10.1136/openhrt-2019-001042}.
\newblock URL \url{https://doi.org/10.1136/openhrt-2019-001042}.

\bibitem[Vedula et~al.(2015)Vedula, George, Younes, and Mittal]{Vedula2015}
Vijay Vedula, Richard George, Laurent Younes, and Rajat Mittal.
\newblock Hemodynamics in the left atrium and its effect on ventricular flow patterns.
\newblock \emph{Journal of Biomechanical Engineering}, 137\penalty0 (11), September 2015.
\newblock \doi{10.1115/1.4031487}.
\newblock URL \url{https://doi.org/10.1115/1.4031487}.

\bibitem[Huang et~al.(2008)Huang, Huang, Huang, Huang, and Huang]{Huang2008}
Xinsheng Huang, Yigao Huang, Tao Huang, Wenhui Huang, and Zhendong Huang.
\newblock Individual pulmonary vein imaging by transthoracic echocardiography: an inadequate traditional interpretation.
\newblock \emph{European Heart Journal - Cardiovascular Imaging}, 9\penalty0 (5):\penalty0 655--660, February 2008.
\newblock \doi{10.1093/ejechocard/jen032}.
\newblock URL \url{https://doi.org/10.1093/ejechocard/jen032}.

\bibitem[Bissell et~al.(2023)Bissell, Raimondi, Ali, Allen, Barker, Bolger, Burris, Carh\"{a}ll, Collins, Ebbers, Francois, Frydrychowicz, Garg, Geiger, Ha, Hennemuth, Hope, Hsiao, Johnson, Kozerke, Ma, Markl, Martins, Messina, Oechtering, van Ooij, Rigsby, Rodriguez-Palomares, Roest, Rold{\'{a}}n-Alzate, Schnell, Sotelo, Stuber, Syed, T\"{o}ger, van~der Geest, Westenberg, Zhong, Zhong, Wieben, and Dyverfeldt]{Bissell2023}
Malenka~M. Bissell, Francesca Raimondi, Lamia~Ait Ali, Bradley~D. Allen, Alex~J. Barker, Ann Bolger, Nicholas Burris, Carl-Johan Carh\"{a}ll, Jeremy~D. Collins, Tino Ebbers, Christopher~J. Francois, Alex Frydrychowicz, Pankaj Garg, Julia Geiger, Hojin Ha, Anja Hennemuth, Michael~D. Hope, Albert Hsiao, Kevin Johnson, Sebastian Kozerke, Liliana~E. Ma, Michael Markl, Duarte Martins, Marci Messina, Thekla~H. Oechtering, Pim van Ooij, Cynthia Rigsby, Jose Rodriguez-Palomares, Arno A.~W. Roest, Alejandro Rold{\'{a}}n-Alzate, Susanne Schnell, Julio Sotelo, Matthias Stuber, Ali~B. Syed, Johannes T\"{o}ger, Rob van~der Geest, Jos Westenberg, Liang Zhong, Yumin Zhong, Oliver Wieben, and Petter Dyverfeldt.
\newblock {4D} flow cardiovascular magnetic resonance consensus statement: 2023 update.
\newblock \emph{Journal of Cardiovascular Magnetic Resonance}, 25\penalty0 (1), July 2023.
\newblock \doi{10.1186/s12968-023-00942-z}.
\newblock URL \url{https://doi.org/10.1186/s12968-023-00942-z}.

\bibitem[Kr\"{a}uter et~al.(2020)Kr\"{a}uter, Reiter, Reiter, Nizhnikava, Masana, Schmidt, Fuchsj\"{a}ger, Stollberger, and Reiter]{Kruter2020}
Corina Kr\"{a}uter, Ursula Reiter, Clemens Reiter, Volha Nizhnikava, Marc Masana, Albrecht Schmidt, Michael Fuchsj\"{a}ger, Rudolf Stollberger, and Gert Reiter.
\newblock Automated mitral valve vortex ring extraction from {4D} flow {MRI}.
\newblock \emph{Magnetic Resonance in Medicine}, 84\penalty0 (6):\penalty0 3396--3408, June 2020.
\newblock \doi{10.1002/mrm.28361}.
\newblock URL \url{https://doi.org/10.1002/mrm.28361}.

\bibitem[Blume et~al.(2011)Blume, Mcleod, Barnes, Seward, Pellikka, Bastiansen, and Tsang]{Blume2011}
G.~G. Blume, C.~J. Mcleod, M.~E. Barnes, J.~B. Seward, P.~A. Pellikka, P.~M. Bastiansen, and T.~S.~M. Tsang.
\newblock Left atrial function: physiology, assessment, and clinical implications.
\newblock \emph{European Journal of Echocardiography}, 12\penalty0 (6):\penalty0 421--430, May 2011.
\newblock \doi{10.1093/ejechocard/jeq175}.
\newblock URL \url{https://doi.org/10.1093/ejechocard/jeq175}.

\bibitem[K\"{o}hler et~al.(2019)K\"{o}hler, Grothoff, Gutberlet, and Preim]{Khler2019}
Benjamin K\"{o}hler, Matthias Grothoff, Matthias Gutberlet, and Bernhard Preim.
\newblock Bloodline: A system for the guided analysis of cardiac {4D} {PC}-{MRI} data.
\newblock \emph{Computers \& Graphics}, 82:\penalty0 32--43, August 2019.
\newblock \doi{10.1016/j.cag.2019.05.004}.
\newblock URL \url{https://doi.org/10.1016/j.cag.2019.05.004}.

\bibitem[Fluckiger et~al.(2013)Fluckiger, Goldberger, Lee, Ng, Lee, Goyal, and Markl]{Fluckiger2013}
Jacob~U. Fluckiger, Jeffrey~J. Goldberger, Daniel~C. Lee, Jason Ng, Richard Lee, Amita Goyal, and Michael Markl.
\newblock Left atrial flow velocity distribution and flow coherence using four-dimensional {FLOW} {MRI}: A pilot study investigating the impact of age and pre- and postintervention atrial fibrillation on atrial hemodynamics.
\newblock \emph{Journal of Magnetic Resonance Imaging}, 38\penalty0 (3):\penalty0 580--587, January 2013.
\newblock \doi{10.1002/jmri.23994}.
\newblock URL \url{https://doi.org/10.1002/jmri.23994}.

\bibitem[Markl et~al.(2016)Markl, Lee, Furiasse, Carr, Foucar, Ng, Carr, and Goldberger]{Markl2016}
Michael Markl, Daniel~C. Lee, Nicholas Furiasse, Maria Carr, Charles Foucar, Jason Ng, James Carr, and Jeffrey~J. Goldberger.
\newblock Left atrial and left atrial appendage {4D} blood flow dynamics in atrial fibrillation.
\newblock \emph{Circulation: Cardiovascular Imaging}, 9\penalty0 (9), September 2016.
\newblock \doi{10.1161/circimaging.116.004984}.
\newblock URL \url{https://doi.org/10.1161/circimaging.116.004984}.

\bibitem[Gaeta et~al.(2018)Gaeta, Dyverfeldt, Eriksson, Carlh\"{a}ll, Ebbers, and Bolger]{Gaeta2018}
Stephen Gaeta, Petter Dyverfeldt, Jonatan Eriksson, Carl-Johan Carlh\"{a}ll, Tino Ebbers, and Ann~F. Bolger.
\newblock Fixed volume particle trace emission for the analysis of left atrial blood flow using {4D} {Flow} {MRI}.
\newblock \emph{Magnetic Resonance Imaging}, 47:\penalty0 83–88, April 2018.
\newblock ISSN 0730-725X.
\newblock \doi{10.1016/j.mri.2017.12.008}.
\newblock URL \url{http://dx.doi.org/10.1016/j.mri.2017.12.008}.

\bibitem[Garcia et~al.(2019)Garcia, Sheitt, Bristow, Lydell, Howarth, Heydari, Prato, Drangova, Thornhill, Nery, Wilton, Skanes, and White]{Garcia2019}
Julio Garcia, Hana Sheitt, Michael~S. Bristow, Carmen Lydell, Andrew~G. Howarth, Bobak Heydari, Frank~S. Prato, Maria Drangova, Rebecca~E. Thornhill, Pablo Nery, Stephen~B. Wilton, Allan Skanes, and James~A. White.
\newblock Left atrial vortex size and velocity distributions by {4D} flow {MRI} in patients with paroxysmal atrial fibrillation: Associations with age and $\text{CHA}_2\text{DS}_2\text{-VASc}$ risk score.
\newblock \emph{Journal of Magnetic Resonance Imaging}, 51\penalty0 (3):\penalty0 871--884, July 2019.
\newblock \doi{10.1002/jmri.26876}.
\newblock URL \url{https://doi.org/10.1002/jmri.26876}.

\bibitem[Demirkiran et~al.(2021)Demirkiran, Amier, Hofman, van~der Geest, Robbers, Hopman, Mulder, van~de Ven, Allaart, van Rossum, G\"{o}tte, and Nijveldt]{Demirkiran2021}
Ahmet Demirkiran, Raquel~P. Amier, Mark B.~M. Hofman, Rob~J. van~der Geest, Lourens F. H.~J. Robbers, Luuk H. G.~A. Hopman, Mark~J. Mulder, Peter van~de Ven, Cornelis~P. Allaart, Albert~C. van Rossum, Marco J.~W. G\"{o}tte, and Robin Nijveldt.
\newblock Altered left atrial {4D} flow characteristics in patients with paroxysmal atrial fibrillation in the absence of apparent remodeling.
\newblock \emph{Scientific Reports}, 11\penalty0 (1), March 2021.
\newblock \doi{10.1038/s41598-021-85176-8}.
\newblock URL \url{https://doi.org/10.1038/s41598-021-85176-8}.

\bibitem[Spartera et~al.(2021)Spartera, Pessoa-Amorim, Stracquadanio, Ende, Fletcher, Manley, Neubauer, Ferreira, Casadei, Hess, and Wijesurendra]{Spartera2021}
Marco Spartera, Guilherme Pessoa-Amorim, Antonio Stracquadanio, Adam~Von Ende, Alison Fletcher, Peter Manley, Stefan Neubauer, Vanessa~M. Ferreira, Barbara Casadei, Aaron~T. Hess, and Rohan~S. Wijesurendra.
\newblock Left atrial {4D} flow cardiovascular magnetic resonance: a reproducibility study in sinus rhythm and atrial fibrillation.
\newblock \emph{Journal of Cardiovascular Magnetic Resonance}, 23\penalty0 (1), March 2021.
\newblock \doi{10.1186/s12968-021-00729-0}.
\newblock URL \url{https://doi.org/10.1186/s12968-021-00729-0}.

\bibitem[Nallamothu et~al.(2024)Nallamothu, Pradella, Markl, Greenland, Passman, and Elbaz]{Nallamothu2024}
Thara Nallamothu, Maurice Pradella, Michael Markl, Philip Greenland, Rod Passman, and Mohammed~SM Elbaz.
\newblock Robust and fast stochastic 4d flow vector-field signature technique for quantifying composite flow dynamics from 4d flow mri: Application to left atrial flow in atrial fibrillation.
\newblock \emph{Medical Image Analysis}, 92:\penalty0 103065, February 2024.
\newblock ISSN 1361-8415.
\newblock \doi{10.1016/j.media.2023.103065}.
\newblock URL \url{http://dx.doi.org/10.1016/j.media.2023.103065}.

\bibitem[Ferdian et~al.(2020)Ferdian, Suinesiaputra, Dubowitz, Zhao, Wang, Cowan, and Young]{4dflownet}
Edward Ferdian, Avan Suinesiaputra, David~J. Dubowitz, Debbie Zhao, Alan Wang, Brett Cowan, and Alistair~A. Young.
\newblock {4DFlowNet}: Super-resolution {4D} flow {MRI} using deep learning and computational fluid dynamics.
\newblock \emph{Frontiers in Physics}, 8, 2020.
\newblock ISSN 2296-424X.
\newblock \doi{10.3389/fphy.2020.00138}.
\newblock URL \url{https://www.frontiersin.org/articles/10.3389/fphy.2020.00138}.

\bibitem[Bustamante et~al.(2017)Bustamante, Gupta, Carlh\"{a}ll, and Ebbers]{Bustamante2017}
Mariana Bustamante, Vikas Gupta, Carl-Johan Carlh\"{a}ll, and Tino Ebbers.
\newblock Improving visualization of {4D} flow cardiovascular magnetic resonance with four-dimensional angiographic data: generation of a {4D} phase-contrast magnetic resonance {CardioAngiography} ({4D} {PC}-{MRCA}).
\newblock \emph{Journal of Cardiovascular Magnetic Resonance}, 19\penalty0 (1), June 2017.
\newblock \doi{10.1186/s12968-017-0360-8}.
\newblock URL \url{https://doi.org/10.1186/s12968-017-0360-8}.

\bibitem[Isensee et~al.(2020)Isensee, Jaeger, Kohl, Petersen, and Maier-Hein]{nnunet}
Fabian Isensee, Paul~F. Jaeger, Simon A.~A. Kohl, Jens Petersen, and Klaus~H. Maier-Hein.
\newblock {nnU}-{Net}: a self-configuring method for deep learning-based biomedical image segmentation.
\newblock \emph{Nature Methods}, 18\penalty0 (2):\penalty0 203--211, December 2020.
\newblock \doi{10.1038/s41592-020-01008-z}.
\newblock URL \url{https://doi.org/10.1038/s41592-020-01008-z}.

\bibitem[Nagueh et~al.(2016)Nagueh, Smiseth, Appleton, Byrd, Dokainish, Edvardsen, Flachskampf, Gillebert, Klein, Lancellotti, Marino, Oh, Popescu, and Waggoner]{Nagueh2016}
Sherif~F. Nagueh, Otto~A. Smiseth, Christopher~P. Appleton, Benjamin~F. Byrd, Hisham Dokainish, Thor Edvardsen, Frank~A. Flachskampf, Thierry~C. Gillebert, Allan~L. Klein, Patrizio Lancellotti, Paolo Marino, Jae~K. Oh, Bogdan~Alexandru Popescu, and Alan~D. Waggoner.
\newblock Recommendations for the evaluation of left ventricular diastolic function by echocardiography: An update from the american society of echocardiography and the european association of cardiovascular imaging.
\newblock \emph{Journal of the American Society of Echocardiography}, 29\penalty0 (4):\penalty0 277--314, April 2016.
\newblock \doi{10.1016/j.echo.2016.01.011}.
\newblock URL \url{https://doi.org/10.1016/j.echo.2016.01.011}.

\bibitem[Cherry et~al.(2022)Cherry, Khatir, Khan, and Bissell]{Cherry2022}
Molly Cherry, Zinedine Khatir, Amirul Khan, and Malenka Bissell.
\newblock The impact of {4D} flow {MRI} spatial resolution on patient-specific {CFD} simulations of the thoracic aorta.
\newblock \emph{Scientific Reports}, 12\penalty0 (1), September 2022.
\newblock \doi{10.1038/s41598-022-19347-6}.
\newblock URL \url{https://doi.org/10.1038/s41598-022-19347-6}.

\bibitem[Marin-Castrillon et~al.(2023)Marin-Castrillon, Geronzi, Boucher, Lin, Morgant, Cochet, Rochette, Leclerc, Ambarki, Jin, Aho, Lalande, Bouchot, and Presles]{MarinCastrillon2023}
Diana~M. Marin-Castrillon, Leonardo Geronzi, Arnaud Boucher, Siyu Lin, Marie-Catherine Morgant, Alexandre Cochet, Michel Rochette, Sarah Leclerc, Khalid Ambarki, Ning Jin, Ludwig~Serge Aho, Alain Lalande, Olivier Bouchot, and Benoit Presles.
\newblock Segmentation of the aorta in systolic phase from {4D} flow {MRI}: multi-atlas vs. deep learning.
\newblock \emph{Magnetic Resonance Materials in Physics, Biology and Medicine}, 36\penalty0 (5):\penalty0 687--700, February 2023.
\newblock \doi{10.1007/s10334-023-01066-2}.
\newblock URL \url{https://doi.org/10.1007/s10334-023-01066-2}.

\bibitem[Garrido-Oliver et~al.(2022)Garrido-Oliver, Aviles, Córdova, Dux-Santoy, Ruiz-Muñoz, Teixido-Tura, Maso~Talou, Morales~Ferez, Jiménez, Evangelista, Ferreira-González, Rodriguez-Palomares, Camara, and Guala]{GarridoOliver2022}
Juan Garrido-Oliver, Jordina Aviles, Marcos~Mejía Córdova, Lydia Dux-Santoy, Aroa Ruiz-Muñoz, Gisela Teixido-Tura, Gonzalo~D. Maso~Talou, Xabier Morales~Ferez, Guillermo Jiménez, Arturo Evangelista, Ignacio Ferreira-González, Jose Rodriguez-Palomares, Oscar Camara, and Andrea Guala.
\newblock Machine learning for the automatic assessment of aortic rotational flow and wall shear stress from {4D} flow cardiac magnetic resonance imaging.
\newblock \emph{European Radiology}, 32\penalty0 (10):\penalty0 7117–7127, August 2022.
\newblock ISSN 1432-1084.
\newblock \doi{10.1007/s00330-022-09068-9}.
\newblock URL \url{http://dx.doi.org/10.1007/s00330-022-09068-9}.

\bibitem[Khan et~al.(2019)Khan, Yang, Zhan, Judd, Chan, Nabi, Heitner, Kim, Klem, Nagueh, and Shah]{Khan2019}
Mohammad~A. Khan, Eric~Y. Yang, Yang Zhan, Robert~M. Judd, Wenyaw Chan, Faisal Nabi, John~F. Heitner, Raymond~J. Kim, Igor Klem, Sherif~F. Nagueh, and Dipan~J. Shah.
\newblock Association of left atrial volume index and all-cause mortality in patients referred for routine cardiovascular magnetic resonance: a multicenter study.
\newblock \emph{Journal of Cardiovascular Magnetic Resonance}, 21\penalty0 (1), January 2019.
\newblock ISSN 1532-429X.
\newblock \doi{10.1186/s12968-018-0517-0}.
\newblock URL \url{http://dx.doi.org/10.1186/s12968-018-0517-0}.

\bibitem[Thomas et~al.(2023)Thomas, Negishi, and Pathan]{Thomas2023}
Liza Thomas, Kazuaki Negishi, and Faraz~Khalid Pathan.
\newblock Editorial: Evaluation of the left atrium: Its role in atrial fibrillation and diastolic function.
\newblock \emph{Frontiers in Cardiovascular Medicine}, 10, February 2023.
\newblock \doi{10.3389/fcvm.2023.1130531}.
\newblock URL \url{https://doi.org/10.3389/fcvm.2023.1130531}.

\bibitem[Mor-Avi et~al.(2012)Mor-Avi, Yodwut, Jenkins, K\"{u}hl, Nesser, Marwick, Franke, Weinert, Niel, Steringer-Mascherbauer, Freed, Sugeng, and Lang]{MorAvi2012}
Victor Mor-Avi, Chattanong Yodwut, Carly Jenkins, Harald K\"{u}hl, Hans-Joachim Nesser, Thomas~H. Marwick, Andreas Franke, Lynn Weinert, Johannes Niel, Regina Steringer-Mascherbauer, Benjamin~H. Freed, Lissa Sugeng, and Roberto~M. Lang.
\newblock Real-time {3D} echocardiographic quantification of left atrial volume.
\newblock \emph{{JACC}: Cardiovascular Imaging}, 5\penalty0 (8):\penalty0 769--777, August 2012.
\newblock \doi{10.1016/j.jcmg.2012.05.011}.
\newblock URL \url{https://doi.org/10.1016/j.jcmg.2012.05.011}.

\bibitem[Nacif et~al.(2012)Nacif, Barranhas, T\"{u}rkbey, Marchiori, Kawel, Mello, Falcão, Junior, and Rochitte]{Nacif2012}
Marcelo~Souto Nacif, Adriana~Dias Barranhas, Evrim T\"{u}rkbey, Edson Marchiori, Nadine Kawel, Ricardo A.~F. Mello, Ricardo~Oliveira Falcão, Amarino C.~Oliveira Junior, and Carlos~Eduardo Rochitte.
\newblock Left atrial volume quantification using cardiac {MRI} in atrial fibrillation: comparison of the simpson’s method with biplane area-length, ellipse, and three-dimensional methods.
\newblock \emph{Diagnostic and Interventional Radiology}, December 2012.
\newblock ISSN 1305-3612.
\newblock \doi{10.5152/dir.2012.002}.
\newblock URL \url{http://dx.doi.org/10.5152/dir.2012.002}.

\bibitem[Crandon et~al.(2017)Crandon, Elbaz, Westenberg, van~der Geest, Plein, and Garg]{Crandon2017}
Saul Crandon, Mohammed~S.M. Elbaz, Jos~J.M. Westenberg, Rob~J. van~der Geest, Sven Plein, and Pankaj Garg.
\newblock Clinical applications of intra-cardiac four-dimensional flow cardiovascular magnetic resonance: A systematic review.
\newblock \emph{International Journal of Cardiology}, 249:\penalty0 486--493, December 2017.
\newblock \doi{10.1016/j.ijcard.2017.07.023}.
\newblock URL \url{https://doi.org/10.1016/j.ijcard.2017.07.023}.

\bibitem[Gupta et~al.(2021)Gupta, Soulat, Avery, Allen, Collins, Choudhury, Bonow, Carr, Markl, and Elbaz]{Gupta2021}
Aakash~N. Gupta, Gilles Soulat, Ryan Avery, Bradley~D. Allen, Jeremy~D. Collins, Lubna Choudhury, Robert~O. Bonow, James Carr, Michael Markl, and Mohammed S.~M. Elbaz.
\newblock {4D} flow {MRI} left atrial kinetic energy in hypertrophic cardiomyopathy is associated with mitral regurgitation and left ventricular outflow tract obstruction.
\newblock \emph{The International Journal of Cardiovascular Imaging}, 37\penalty0 (9):\penalty0 2755--2765, February 2021.
\newblock \doi{10.1007/s10554-021-02167-6}.
\newblock URL \url{https://doi.org/10.1007/s10554-021-02167-6}.

\bibitem[Barker et~al.(2013)Barker, van Ooij, Bandi, Garcia, Albaghdadi, McCarthy, Bonow, Carr, Collins, Malaisrie, and Markl]{Barker2013}
Alex~J. Barker, Pim van Ooij, Krishna Bandi, Julio Garcia, Mazen Albaghdadi, Patrick McCarthy, Robert~O. Bonow, James Carr, Jeremy Collins, S.~Chris Malaisrie, and Michael Markl.
\newblock Viscous energy loss in the presence of abnormal aortic flow.
\newblock \emph{Magnetic Resonance in Medicine}, 72\penalty0 (3):\penalty0 620--628, October 2013.
\newblock \doi{10.1002/mrm.24962}.
\newblock URL \url{https://doi.org/10.1002/mrm.24962}.

\bibitem[Sch\"{a}fer et~al.(2017)Sch\"{a}fer, Humphries, Stenmark, Kheyfets, Buckner, Hunter, and Fenster]{Schfer2017}
Michal Sch\"{a}fer, Stephen Humphries, Kurt~R Stenmark, Vitaly~O Kheyfets, J~Kern Buckner, Kendall~S Hunter, and Brett~E Fenster.
\newblock {4D} flow cardiac magnetic resonance-derived vorticity is sensitive marker of left ventricular diastolic dysfunction in patients with mild-to-moderate chronic obstructive pulmonary disease.
\newblock \emph{European Heart Journal - Cardiovascular Imaging}, 19\penalty0 (4):\penalty0 415--424, April 2017.
\newblock \doi{10.1093/ehjci/jex069}.
\newblock URL \url{https://doi.org/10.1093/ehjci/jex069}.

\bibitem[G\"{u}nther and Theisel(2018)]{Gnther2018}
Tobias G\"{u}nther and Holger Theisel.
\newblock The state of the art in vortex extraction.
\newblock \emph{Computer Graphics Forum}, 37\penalty0 (6):\penalty0 149--173, January 2018.
\newblock \doi{10.1111/cgf.13319}.
\newblock URL \url{https://doi.org/10.1111/cgf.13319}.

\bibitem[Hunt et~al.(1988)Hunt, Wray, and Moin]{hunt1988eddies}
Julian~CR Hunt, Alan~A Wray, and Parviz Moin.
\newblock Eddies, streams, and convergence zones in turbulent flows.
\newblock \emph{Studying turbulence using numerical simulation databases, 2. Proceedings of the 1988 summer program}, 1988.

\bibitem[Marlevi et~al.(2021{\natexlab{a}})Marlevi, Balmus, Hessenthaler, Viola, Fovargue, de~Vecchi, Lamata, Burris, Pagani, Engvall, Edelman, Ebbers, and Nordsletten]{Marlevi2021}
David Marlevi, Maximilian Balmus, Andreas Hessenthaler, Federica Viola, Daniel Fovargue, Adelaide de~Vecchi, Pablo Lamata, Nicholas~S. Burris, Francis~D. Pagani, Jan Engvall, Elazer~R. Edelman, Tino Ebbers, and David~A. Nordsletten.
\newblock Non-invasive estimation of relative pressure for intracardiac flows using virtual work-energy.
\newblock \emph{Medical Image Analysis}, 68:\penalty0 101948, February 2021{\natexlab{a}}.
\newblock \doi{10.1016/j.media.2020.101948}.
\newblock URL \url{https://doi.org/10.1016/j.media.2020.101948}.

\bibitem[Marlevi et~al.(2019)Marlevi, Ruijsink, Balmus, Dillon-Murphy, Fovargue, Pushparajah, Bertoglio, Colarieti-Tosti, Larsson, Lamata, Figueroa, Razavi, and Nordsletten]{Marlevi2019}
David Marlevi, Bram Ruijsink, Maximilian Balmus, Desmond Dillon-Murphy, Daniel Fovargue, Kuberan Pushparajah, Crist{\'{o}}bal Bertoglio, Massimiliano Colarieti-Tosti, Matilda Larsson, Pablo Lamata, C.~Alberto Figueroa, Reza Razavi, and David~A. Nordsletten.
\newblock Estimation of cardiovascular relative pressure using virtual work-energy.
\newblock \emph{Scientific Reports}, 9\penalty0 (1), February 2019.
\newblock \doi{10.1038/s41598-018-37714-0}.
\newblock URL \url{https://doi.org/10.1038/s41598-018-37714-0}.

\bibitem[Vos et~al.(2023)Vos, Raafs, Henkens, Pedrizzetti, van Deursen, Rodwell, Heymans, and Nijveldt]{Vos2023}
Jacqueline~L Vos, Anne~G Raafs, Michiel T H~M Henkens, Gianni Pedrizzetti, Caroline~J van Deursen, Laura Rodwell, Stephane R~B Heymans, and Robin Nijveldt.
\newblock {CMR}-derived left ventricular intraventricular pressure gradients identify different patterns associated with prognosis in dilated cardiomyopathy.
\newblock \emph{European Heart Journal - Cardiovascular Imaging}, 24\penalty0 (9):\penalty0 1231–1240, May 2023.
\newblock ISSN 2047-2412.
\newblock \doi{10.1093/ehjci/jead083}.
\newblock URL \url{http://dx.doi.org/10.1093/ehjci/jead083}.

\bibitem[Sotelo et~al.(2019)Sotelo, Mura, Hurtado, and Uribe]{sotelo2019novel}
Julio Sotelo, Joaquin Mura, Daniel Hurtado, and Sergio Uribe.
\newblock A novel matlab toolbox for processing 4d flow mri data.
\newblock In \emph{Proc. Intl. Soc. Mag. Reson. Med}, volume~27, page 1956, 2019.

\bibitem[Heiberg et~al.(2012)Heiberg, Green, Toger, Andersson, Carlsson, and Arheden]{Heiberg2012}
Einar Heiberg, Christopher Green, Johannes Toger, Andreas~M Andersson, Marcus Carlsson, and Hakan Arheden.
\newblock {FourFlow} - open source code software for quantification and visualization of time-resolved three-directional phase contrast magnetic resonance velocity mapping.
\newblock \emph{Journal of Cardiovascular Magnetic Resonance}, 14\penalty0 (S1), February 2012.
\newblock ISSN 1532-429X.
\newblock \doi{10.1186/1532-429x-14-s1-w14}.
\newblock URL \url{http://dx.doi.org/10.1186/1532-429X-14-S1-W14}.

\bibitem[Heiberg et~al.(2010)Heiberg, Sj\"{o}gren, Ugander, Carlsson, Engblom, and Arheden]{Heiberg2010}
Einar Heiberg, Jane Sj\"{o}gren, Martin Ugander, Marcus Carlsson, Henrik Engblom, and Håkan Arheden.
\newblock Design and validation of segment - freely available software for cardiovascular image analysis.
\newblock \emph{BMC Medical Imaging}, 10\penalty0 (1), January 2010.
\newblock ISSN 1471-2342.
\newblock \doi{10.1186/1471-2342-10-1}.
\newblock URL \url{http://dx.doi.org/10.1186/1471-2342-10-1}.

\bibitem[Barrera-Naranjo et~al.(2023)Barrera-Naranjo, Marin-Castrillon, Decourselle, Lin, Leclerc, Morgant, Bernard, Oliveira, Boucher, Presles, Bouchot, Christophe, and Lalande]{BarreraNaranjo2023}
Armando Barrera-Naranjo, Diana~M. Marin-Castrillon, Thomas Decourselle, Siyu Lin, Sarah Leclerc, Marie-Catherine Morgant, Chlo{\'{e}} Bernard, Shirley~De Oliveira, Arnaud Boucher, Benoit Presles, Olivier Bouchot, Jean-Joseph Christophe, and Alain Lalande.
\newblock Segmentation of {4D} flow {MRI}: Comparison between {3D} deep learning and velocity-based level sets.
\newblock \emph{Journal of Imaging}, 9\penalty0 (6):\penalty0 123, June 2023.
\newblock \doi{10.3390/jimaging9060123}.
\newblock URL \url{https://doi.org/10.3390/jimaging9060123}.

\bibitem[Bustamante et~al.(2018)Bustamante, Gupta, Forsberg, Carlh\"{a}ll, Engvall, and Ebbers]{Bustamante2018}
Mariana Bustamante, Vikas Gupta, Daniel Forsberg, Carl-Johan Carlh\"{a}ll, Jan Engvall, and Tino Ebbers.
\newblock Automated multi-atlas segmentation of cardiac {4D} flow {MRI}.
\newblock \emph{Medical Image Analysis}, 49:\penalty0 128--140, October 2018.
\newblock \doi{10.1016/j.media.2018.08.003}.
\newblock URL \url{https://doi.org/10.1016/j.media.2018.08.003}.

\bibitem[Maroun et~al.(2023)Maroun, Baraboo, Gunasekaran, Hwang, Liu, Passman, Kim, Allen, Markl, and Pradella]{Maroun2023}
Anthony Maroun, Justin~J. Baraboo, Suvai Gunasekaran, Julia~M. Hwang, Sophia~Z. Liu, Rod~S. Passman, Daniel Kim, Bradley~D. Allen, Michael Markl, and Maurice Pradella.
\newblock Comparison of biplane area-length method and {3D} volume quantification by using cardiac {MRI} for assessment of left atrial volume in atrial fibrillation.
\newblock \emph{Radiology: Cardiothoracic Imaging}, 5\penalty0 (2), April 2023.
\newblock ISSN 2638-6135.
\newblock \doi{10.1148/ryct.220133}.
\newblock URL \url{http://dx.doi.org/10.1148/ryct.220133}.

\bibitem[Dyverfeldt et~al.(2015)Dyverfeldt, Bissell, Barker, Bolger, Carlh\"{a}ll, Ebbers, Francios, Frydrychowicz, Geiger, Giese, Hope, Kilner, Kozerke, Myerson, Neubauer, Wieben, and Markl]{Dyverfeldt2015}
Petter Dyverfeldt, Malenka Bissell, Alex~J. Barker, Ann~F. Bolger, Carl-Johan Carlh\"{a}ll, Tino Ebbers, Christopher~J. Francios, Alex Frydrychowicz, Julia Geiger, Daniel Giese, Michael~D. Hope, Philip~J. Kilner, Sebastian Kozerke, Saul Myerson, Stefan Neubauer, Oliver Wieben, and Michael Markl.
\newblock {4D} flow cardiovascular magnetic resonance consensus statement.
\newblock \emph{Journal of Cardiovascular Magnetic Resonance}, 17\penalty0 (1), August 2015.
\newblock \doi{10.1186/s12968-015-0174-5}.
\newblock URL \url{https://doi.org/10.1186/s12968-015-0174-5}.

\bibitem[Alattar et~al.(2022)Alattar, Soulat, Gencer, Messas, Bollache, Kachenoura, and Mousseaux]{Alattar2022}
Yousef Alattar, Gilles Soulat, Umit Gencer, Emmanuel Messas, Emilie Bollache, Nadjia Kachenoura, and Elie Mousseaux.
\newblock Left ventricular diastolic early and late filling quantified from {4D} flow magnetic resonance imaging.
\newblock \emph{Diagnostic and Interventional Imaging}, 103\penalty0 (7-8):\penalty0 345--352, July 2022.
\newblock \doi{10.1016/j.diii.2022.02.003}.
\newblock URL \url{https://doi.org/10.1016/j.diii.2022.02.003}.

\bibitem[Wieslander et~al.(2019)Wieslander, Ramos, Ax, Petersson, and Ugander]{Wieslander2019}
Bj\"{o}rn Wieslander, Joao~G{\'{e}}nio Ramos, Malin Ax, Johan Petersson, and Martin Ugander.
\newblock Supine, prone, right and left gravitational effects on human pulmonary circulation.
\newblock \emph{Journal of Cardiovascular Magnetic Resonance}, 21\penalty0 (1), November 2019.
\newblock \doi{10.1186/s12968-019-0577-9}.
\newblock URL \url{https://doi.org/10.1186/s12968-019-0577-9}.

\bibitem[Arvidsson et~al.(2013)Arvidsson, T\"{o}ger, Heiberg, Carlsson, and Arheden]{Arvidsson2013}
Per~M. Arvidsson, Johannes T\"{o}ger, Einar Heiberg, Marcus Carlsson, and H{\aa}kan Arheden.
\newblock Quantification of left and right atrial kinetic energy using four-dimensional intracardiac magnetic resonance imaging flow measurements.
\newblock \emph{Journal of Applied Physiology}, 114\penalty0 (10):\penalty0 1472--1481, May 2013.
\newblock \doi{10.1152/japplphysiol.00932.2012}.
\newblock URL \url{https://doi.org/10.1152/japplphysiol.00932.2012}.

\bibitem[Suwa et~al.(2014)Suwa, Saitoh, Takehara, Sano, Nobuhara, Saotome, Urushida, Katoh, Satoh, Sugiyama, Wakayama, Alley, Sakahara, and Hayashi]{Suwa2014}
Kenichiro Suwa, Takeji Saitoh, Yasuo Takehara, Makoto Sano, Mamoru Nobuhara, Masao Saotome, Tsuyoshi Urushida, Hideki Katoh, Hiroshi Satoh, Masataka Sugiyama, Tetsuya Wakayama, Marcus Alley, Harumi Sakahara, and Hideharu Hayashi.
\newblock Characteristics of intra-left atrial flow dynamics and factors affecting formation of the vortex flow.
\newblock \emph{Circulation Journal}, 79\penalty0 (1):\penalty0 144--152, 2014.
\newblock \doi{10.1253/circj.cj-14-0562}.
\newblock URL \url{https://doi.org/10.1253/circj.cj-14-0562}.

\bibitem[Sekine et~al.(2022)Sekine, Nakaza, Matsumoto, Ando, Inoue, Sakamoto, Maruyama, Obara, Leonowicz, Usuda, and Kumita]{Sekine2022}
Tetsuro Sekine, Masatoki Nakaza, Mitsuo Matsumoto, Takahiro Ando, Tatsuya Inoue, Shun-Ichiro Sakamoto, Mitsunori Maruyama, Makoto Obara, Olgierd Leonowicz, Jitsuo Usuda, and Shinichiro Kumita.
\newblock {4D} flow {MR} imaging of the left atrium: What is non-physiological blood flow in the cardiac system?
\newblock \emph{Magnetic Resonance in Medical Sciences}, 21\penalty0 (2):\penalty0 293--308, 2022.
\newblock \doi{10.2463/mrms.rev.2021-0137}.
\newblock URL \url{https://doi.org/10.2463/mrms.rev.2021-0137}.

\bibitem[Spartera et~al.(2023)Spartera, Stracquadanio, Pessoa-Amorim, Harston, Mazzucco, Young, Ende, Hess, Ferreira, Kennedy, Neubauer, Casadei, and Wijesurendra]{Spartera_2023}
Marco Spartera, Antonio Stracquadanio, Guilherme Pessoa-Amorim, George Harston, Sara Mazzucco, Victoria Young, Adam~Von Ende, Aaron~T. Hess, Vanessa~M. Ferreira, James Kennedy, Stefan Neubauer, Barbara Casadei, and Rohan~S. Wijesurendra.
\newblock Reduced left atrial rotational flow is independently associated with embolic brain infarcts.
\newblock \emph{{JACC}: Cardiovascular Imaging}, 16\penalty0 (9):\penalty0 1149--1159, sep 2023.
\newblock \doi{10.1016/j.jcmg.2023.03.006}.
\newblock URL \url{https://doi.org/10.1016%2Fj.jcmg.2023.03.006}.

\bibitem[Arvidsson et~al.(2017)Arvidsson, T\"{o}ger, Carlsson, Steding-Ehrenborg, Pedrizzetti, Heiberg, and Arheden]{Arvidsson2017}
Per~M. Arvidsson, Johannes T\"{o}ger, Marcus Carlsson, Katarina Steding-Ehrenborg, Gianni Pedrizzetti, Einar Heiberg, and Håkan Arheden.
\newblock Left and right ventricular hemodynamic forces in healthy volunteers and elite athletes assessed with {4D} flow magnetic resonance imaging.
\newblock \emph{American Journal of Physiology-Heart and Circulatory Physiology}, 312\penalty0 (2):\penalty0 H314–H328, February 2017.
\newblock ISSN 1522-1539.
\newblock \doi{10.1152/ajpheart.00583.2016}.
\newblock URL \url{http://dx.doi.org/10.1152/ajpheart.00583.2016}.

\bibitem[Myronenko et~al.(2020)Myronenko, Yang, Buch, Xu, Ihsani, Doyle, Michalski, Tenenholtz, and Roth]{myronenko20204d}
Andriy Myronenko, Dong Yang, Varun Buch, Daguang Xu, Alvin Ihsani, Sean Doyle, Mark Michalski, Neil Tenenholtz, and Holger Roth.
\newblock 4d cnn for semantic segmentation of cardiac volumetric sequences.
\newblock In \emph{Statistical Atlases and Computational Models of the Heart. Multi-Sequence CMR Segmentation, CRT-EPiggy and LV Full Quantification Challenges: 10th International Workshop, STACOM 2019, Held in Conjunction with MICCAI 2019, Shenzhen, China, October 13, 2019, Revised Selected Papers 10}, pages 72--80. Springer, 2020.

\bibitem[Corrado et~al.(2022)Corrado, Wentland, Starekova, Dhyani, Goss, and Wieben]{Corrado2022}
Philip~A. Corrado, Andrew~L. Wentland, Jitka Starekova, Archana Dhyani, Kara~N. Goss, and Oliver Wieben.
\newblock Fully automated intracardiac {4D} flow {MRI} post-processing using deep learning for biventricular segmentation.
\newblock \emph{European Radiology}, 32\penalty0 (8):\penalty0 5669--5678, February 2022.
\newblock \doi{10.1007/s00330-022-08616-7}.
\newblock URL \url{https://doi.org/10.1007/s00330-022-08616-7}.

\bibitem[Juffermans et~al.(2021)Juffermans, Minderhoud, Wittgren, Kilburg, Ese, Fidock, Zheng, Zhang, Blanken, Lamb, Goeman, Carlsson, Zhao, Planken, van Ooij, Zhong, Chen, Garg, Emrich, Hirsch, T\"{o}ger, and Westenberg]{Juffermans2021}
Joe~F. Juffermans, Savine~C.S. Minderhoud, Johan Wittgren, Anton Kilburg, Amir Ese, Benjamin Fidock, Yu-Cong Zheng, Jun-Mei Zhang, Carmen~P.S. Blanken, Hildo~J. Lamb, Jelle~J. Goeman, Marcus Carlsson, Shihua Zhao, R.~Nils Planken, Pim van Ooij, Liang Zhong, Xiuyu Chen, Pankaj Garg, Tilman Emrich, Alexander Hirsch, Johannes T\"{o}ger, and Jos~J.M. Westenberg.
\newblock Multicenter consistency assessment of valvular flow quantification with automated valve tracking in {4D} flow {CMR}.
\newblock \emph{{JACC}: Cardiovascular Imaging}, 14\penalty0 (7):\penalty0 1354--1366, July 2021.
\newblock \doi{10.1016/j.jcmg.2020.12.014}.
\newblock URL \url{https://doi.org/10.1016/j.jcmg.2020.12.014}.

\bibitem[Marlevi et~al.(2021{\natexlab{b}})Marlevi, Sotelo, Grogan-Kaylor, Ahmed, Uribe, Patel, Edelman, Nordsletten, and Burris]{marlevi2021false}
David Marlevi, Julio~A Sotelo, Ross Grogan-Kaylor, Yunus Ahmed, Sergio Uribe, Himanshu~J Patel, Elazer~R Edelman, David~A Nordsletten, and Nicholas~S Burris.
\newblock False lumen pressure estimation in type b aortic dissection using 4d flow cardiovascular magnetic resonance: comparisons with aortic growth.
\newblock \emph{Journal of Cardiovascular Magnetic Resonance}, 23\penalty0 (1):\penalty0 51, 2021{\natexlab{b}}.

\bibitem[Ferdian et~al.(2023)Ferdian, Marlevi, Schollenberger, Aristova, Edelman, Schnell, Figueroa, Nordsletten, and Young]{ferdian2023cerebrovascular}
Edward Ferdian, David Marlevi, Jonas Schollenberger, Maria Aristova, Elazer~R Edelman, Susanne Schnell, C~Alberto Figueroa, DA~Nordsletten, and Alistair~A Young.
\newblock Cerebrovascular super-resolution 4d flow mri--sequential combination of resolution enhancement by deep learning and physics-informed image processing to non-invasively quantify intracranial velocity, flow, and relative pressure.
\newblock \emph{Medical Image Analysis}, 88:\penalty0 102831, 2023.

\bibitem[Albors et~al.(2022)Albors, Mill, Kjeldsberg, Vilad{\'e}s~Medel, Olivares, Valen-Sendstad, and Camara]{albors2022sensitivity}
Carlos Albors, Jordi Mill, Henrik~A Kjeldsberg, David Vilad{\'e}s~Medel, Andy~L Olivares, Kristian Valen-Sendstad, and Oscar Camara.
\newblock Sensitivity analysis of left atrial wall modeling approaches and inlet/outlet boundary conditions in fluid simulations to predict thrombus formation.
\newblock In \emph{International Workshop on Statistical Atlases and Computational Models of the Heart}, pages 179--189. Springer, 2022.

\bibitem[Morales et~al.(2021)Morales, Mill, Delso, Loncaric, Doltra, Freixa, Sitges, Bijnens, and Camara]{morales20214d}
Xabier Morales, Jordi Mill, Gaspar Delso, Filip Loncaric, Ada Doltra, Xavier Freixa, Marta Sitges, Bart Bijnens, and Oscar Camara.
\newblock 4d flow magnetic resonance imaging for left atrial haemodynamic characterization and model calibration.
\newblock In \emph{Statistical Atlases and Computational Models of the Heart. M\&Ms and EMIDEC Challenges: 11th International Workshop, STACOM 2020, Held in Conjunction with MICCAI 2020, Lima, Peru, October 4, 2020, Revised Selected Papers 11}, pages 156--165. Springer, 2021.

\bibitem[van~de Bovenkamp et~al.(2021)van~de Bovenkamp, Enait, de~Man, Oosterveer, Bogaard, Noordegraaf, van Rossum, and Handoko]{vandeBovenkamp2021}
Arno~A. van~de Bovenkamp, Vidya Enait, Frances~S. de~Man, Frank T.~P. Oosterveer, Harm~Jan Bogaard, Anton~Vonk Noordegraaf, Albert~C. van Rossum, and M.~Louis Handoko.
\newblock Validation of the 2016 {ASE}/{EACVI} guideline for diastolic dysfunction in patients with unexplained dyspnea and a preserved left ventricular ejection fraction.
\newblock \emph{Journal of the American Heart Association}, 10\penalty0 (18), September 2021.
\newblock \doi{10.1161/jaha.121.021165}.
\newblock URL \url{https://doi.org/10.1161/jaha.121.021165}.

\bibitem[Loncaric et~al.(2021)Loncaric, Marti~Castellote, Sanchez-Martinez, Fabijanovic, Nunno, Mimbrero, Sanchis, Doltra, Montserrat, Cikes, Crispi, Piella, Sitges, and Bijnens]{Loncaric2021}
Filip Loncaric, Pablo-Miki Marti~Castellote, Sergio Sanchez-Martinez, Dora Fabijanovic, Loredana Nunno, Maria Mimbrero, Laura Sanchis, Adelina Doltra, Silvia Montserrat, Maja Cikes, Fatima Crispi, Gema Piella, Marta Sitges, and Bart Bijnens.
\newblock Automated pattern recognition in whole-cardiac cycle echocardiographic data: Capturing functional phenotypes with machine learning.
\newblock \emph{Journal of the American Society of Echocardiography}, 34\penalty0 (11):\penalty0 1170–1183, November 2021.
\newblock ISSN 0894-7317.
\newblock \doi{10.1016/j.echo.2021.06.014}.
\newblock URL \url{http://dx.doi.org/10.1016/j.echo.2021.06.014}.

\end{thebibliography}






\clearpage

\section*{Appendix}

\setcounter{section}{0} \label{sup:Statistical}
\renewcommand{\thesection}{A\arabic{section}}

\setcounter{table}{0}
\renewcommand{\thetable}{A\arabic{table}}

\setcounter{figure}{0}
\renewcommand{\thefigure}{A\arabic{figure}}

\section{Statistical analysis results}\label{statistical}

\vspace{-1em}

\begin{table}[h]
    \begin{minipage}{\textwidth}
    \caption{Average values and standard error of the mean of the kinetic energy (KE) and viscous energy loss (VEL) peaks, and the ratio between them (KE/VEL) for the different pathologies. The ANCOVA P-values and effect size ($\eta_{p}^{2}$) results are shown for the effect of pathology and age as a covariable. The P-values are color-coded, ranging from green (0.0) to red (0.2), indicating the level of statistical significance. Similarly, $\eta_{p}^{2}$ is represented using a color gradient, ranging from white (0.0) to blue (0.4) for pathology and white (0.0) to red (0.4) for age, indicating the strength of the association. Lastly, the posthoc results are shown for the pairwise comparisons between pathologies if they are found to be statistically significant ($\alpha<0.5$).}
    \label{tab:EnergyPeaks}
    \small
    \resizebox{\textwidth}{!}{%
    \begin{tabular}{lllllllllllll}
\hline
 &  & \multicolumn{6}{l}{Pathology} & \multicolumn{2}{l}{ANCOVA:   P-value} & \multicolumn{2}{l}{ANCOVA: $\eta_{p}^{2}$} & Posthoc \\ \hline
 &  & Control & HCM & G1 & G2 & G2 - SAM & Hypertensive & Pathology & Age & Pathology & Age & Tukey and Cohen's D \\ \hline
\multirow{3}{*}{KE$_{LA}$ (J/m$^3$)} & S & 29.11 ± 2.61 & 24.39 ± 2.08 & 21.85 ± 3.89 & 22.1 ± 2.39 & 42.26 ± 4.92 & 25.57 ± 1.29 & \gradient{ .005} & \gradient{ .34} & \gradienteta{ .233} & \gradientetados{ .015} & G1 \textless G2 - SAM; P = .045; D = -2.22, G2 \textless G2 - SAM; P =   .045; D = -2.65, G2 - SAM \textgreater HCM; P = .045; D = 2.42, G2 - SAM \textgreater   Hypertensive; P = .045; D = 2.36 \\ \cline{2-13} 
 & E & 40.9 ± 3.28 & 29.92 ± 5.48 & 19.16 ± 1.57 & 21.43 ± 3.16 & 16.39 ± 3.74 & 23.31 ± 1.5 & \gradient{ .197} & \gradient{ .0} & \gradienteta{ .109} & \gradientetados{ .277} &  \\ \cline{2-13} 
 & A & 18.91 ± 1.63 & 16.81 ± 3.42 & 16.41 ± 2.69 & 13.23 ± 1.65 & 9.44 ± 2.02 & 14.63 ± 1.36 & \gradient{ .006} & \gradient{ .003} & \gradienteta{ .228} & \gradientetados{ .134} &  \\ \hline
\multirow{3}{*}{VEL$_{LA}$  (W/m$^3$)} & S & 2.85 ± 0.22 & 3.43 ± 0.32 & 2.35 ± 0.37 & 4.48 ± 0.7 & 2.6 ± 0.35 & 4.18 ± 0.38 & \gradient{ .016} & \gradient{ .569} & \gradienteta{ .197} & \gradientetados{ .005} &  \\ \cline{2-13} 
 & E & 4.71 ± 0.32 & 3.31 ± 0.41 & 2.15 ± 0.31 & 2.62 ± 0.37 & 1.54 ± 0.35 & 2.78 ± 0.23 & \gradient{ .029} & \gradient{ .002} & \gradienteta{ .178} & \gradientetados{ .139} & G1 \textless N; P = .005; D = -1.93, G2 \textless N; P = .022; D = -1.57, G2 - SAM \textless N; P = .001; D = -2.35, Hypertensive \textless N; P = .0; D = -1.48 \\ \cline{2-13} 
 & A & 3.06 ± 0.25 & 2.54 ± 0.5 & 2.64 ± 0.51 & 2.49 ± 0.46 & 1.5 ± 0.35 & 2.47 ± 0.18 & \gradient{ .002} & \gradient{ .001} & \gradienteta{ .252} & \gradientetados{ .161} &  \\ \hline
\multirow{3}{*}{KE/VEL} & S & 10.47 ± 0.8 & 7.3 ± 0.45 & 9.82 ± 1.63 & 5.4 ± 0.71 & 18.43 ± 5.57 & 7.14 ± 0.56 & \gradient{ .0} & \gradient{ .715} & \gradienteta{ .392} & \gradientetados{ .002} & G1 \textless G2 - SAM; P = .041; D = -1.1, G2 \textless G2 - SAM; P = .0; D   = -1.87, G2 - SAM \textgreater HCM; P = .0; D = 1.88, G2 - SAM \textgreater Hypertensive; P =   .0; D = 2.49, G2 - SAM \textgreater N; P = .013; D = 1.51 \\ \cline{2-13} 
 & E & 8.88 ± 0.67 & 8.82 ± 0.78 & 9.97 ± 2.12 & 8.38 ± 0.83 & 10.76 ± 1.57 & 8.99 ± 0.47 & \gradient{ .407} & \gradient{ .089} & \gradienteta{ .077} & \gradientetados{ .046} &  \\ \cline{2-13} 
 & A & 6.29 ± 0.34 & 6.76 ± 1.0 & 6.5 ± 0.89 & 5.78 ± 0.71 & 6.68 ± 1.0 & 6.07 ± 0.32 & \gradient{ .92} & \gradient{ .728} & \gradienteta{ .022} & \gradientetados{ .002} &  \\ \hline
\end{tabular}%
}
\end{minipage}
\end{table}

\vspace{-1em}

\begin{table}[h]
    \begin{minipage}{\textwidth}
    \caption{Average peak values and standard error of the mean for the vorticity magnitude normalized by the left atrium volume ($\abs{\omega_{LA}}$) and the ratio of voxels with Q-Criterion above 500 s$^{-2}$ (Q-Crit$_{500}$). The ANCOVA P-values and effect size ($\eta_{p}^{2}$) results are shown for the effect of pathology and age as a covariable. The P-values are color-coded, ranging from green (0.0) to red (0.2), indicating the level of statistical significance. Similarly, $\eta_{p}^{2}$ is represented using a color gradient, ranging from white (0.0) to blue (0.4) for Pathology and white (0.0) to red (0.4) for age, indicating the strength of the association. Lastly, the posthoc results are shown for the pairwise comparisons between pathologies if they are found to be statistically significant ($\alpha<0.5$).}
    \label{tab:VorticityPeaks}
    \small
    \resizebox{\textwidth}{!}{%
    \begin{tabular}{lllllllllllll}
\hline
 &  & \multicolumn{6}{l}{Pathology} & \multicolumn{2}{l}{ANCOVA:   P-value} & \multicolumn{2}{l}{ANCOVA: $\eta_{p}^{2}$} & Posthoc \\ \hline
 &  & Control & HCM & G1 & G2 & G2 - SAM & Hypertensive & Pathology & Age & Pathology & Age & Tukey and Cohen's D \\ \hline
\multirow{3}{*}{$\abs{\omega}$ ($s^{-1} m^{-3}$)} & S & 73.35 ± 2.52 & 68.59 ± 3.16 & 61.42 ± 5.0 & 67.27 ± 4.76 & 71.98 ± 3.24 & 68.84 ± 1.85 & \gradient{ .546} & \gradient{ .082} & \gradienteta{ .061} & \gradientetados{ .048} &  \\ \cline{2-13} 
 & E & 73.84 ± 2.48 & 60.89 ± 4.31 & 51.94 ± 3.86 & 53.13 ± 4.38 & 45.24 ± 5.06 & 53.29 ± 1.75 & \gradient{ .007} & \gradient{ .121} & \gradienteta{ .221} & \gradientetados{ .092} & G1 \textless N; P = .004; D = -2.04, G2 \textless N; P = .004; D = -1.88, G2   SAM \textless N; P = .0; D = -2.61, Hypertensive \textless N; P = .0; D = -2.06 \\ \cline{2-13} 
 & A & 53.3 ± 1.99 & 45.7 ± 3.67 & 48.79 ± 4.6 & 42.29 ± 1.27 & 36.44 ± 2.4 & 42.16 ± 1.39 & \gradient{ .0} & \gradient{ .094} & \gradienteta{ .312} & \gradientetados{ .044} & G2 - SAM \textless N; P = .037; D = -1.99, Hypertensive \textless N; P = .004; D =   -1.39 \\ \hline
\multirow{3}{*}{Q-crit$_{500}$ (\%)} & S & 14.2 ± 0.8 & 11.1 ± 1.2 & 10.5 ± 1.6 & 9.6 ± 1.0 & 15.9 ± 0.4 & 9.8 ± 0.4 & \gradient{ .0} & \gradient{ .108} & \gradienteta{ .339} & \gradientetados{ .041} & G2 - SAM \textgreater Hypertensive; P = .018; D = 2.83, Hypertensive \textless N; P = .0;   D = -1.59 \\ \cline{2-13} 
 & E & 11.9 ± 0.8 & 6.9 ± 1.1 & 6.2 ± 1.2 & 4.3 ± 0.8 & 6.7 ± 2.2 & 4.9 ± 0.4 & \gradient{ .0} & \gradient{ .167} & \gradienteta{ .369} & \gradientetados{ .085} & G1 \textless N; P = .014; D = -1.59, G2 \textless N; P = .0; D = -2.23, HCM   \textless N; P = .007; D = -1.36, Hypertensive \textless N; P = .0; D = -2.45 \\ \cline{2-13} 
 & A & 6.3 ± 0.5 & 4.4 ± 0.9 & 5.7 ± 1.5 & 3.3 ± 0.3 & 2.9 ± 0.6 & 3.4 ± 0.3 & \gradient{ .001} & \gradient{ .186} & \gradienteta{ .279} & \gradientetados{ .028} & Hypertensive \textless N; P = .006; D = -1.41 \\ \hline
\end{tabular}%
    }
\end{minipage}
\end{table}

\vspace{-1em}

\begin{table}[h]
    \begin{minipage}{\textwidth}
    \caption{Average peak values and standard error of the mean for the vWERP-derived relative pressure (mmHg) measurements. The ANCOVA P-values and effect size ($\eta_{p}^{2}$) results are shown for the effect of pathology and age as a covariable. The P-values are color-coded, ranging from green (0.0) to red (0.2), indicating the level of statistical significance. Similarly, $\eta_{p}^{2}$ is represented using a color gradient, ranging from white (0.0) to blue (0.4) for Pathology and white (0.0) to red (0.4) for age, indicating the strength of the association. Lastly, the posthoc results are shown for the pairwise comparisons between pathologies if they are found to be statistically significant ($\alpha<0.5$).}
    \label{tab:PressurePeaks}
    \small
    \resizebox{\textwidth}{!}{%
    \begin{tabular}{cclllllllllll}
\hline
 &  & \multicolumn{6}{l}{Pathology} & \multicolumn{4}{l}{ANCOVA} & Posthoc \\ \hline
 &  & Control & HCM & G1 & G2 & G2 - SAM & Hypertensive & P-value Pathology & P-value Age & $\eta_{p}^{2}$ Pathology & $\eta_{p}^{2}$ Age & Tukey and Cohen's D \\ \hline
\multirow{4}{*}{$\Delta$P (mmHg)} & $\Delta$E$_{max}$ & 3.2 ± 0.23 & 1.81 ± 0.33 & 1.99 ± 0.4 & 1.35 ± 0.11 & 1.97 ± 0.41 & 1.63 ± 0.09 & \gradient{ .0} & \gradient{ .823} & \gradienteta{ .447} & \gradientetados{ .001} & G2 \textless N; P = .0; D = -2.21, HCM \textless N; P = .007; D = -1.52,   Hypertensive \textless N; P = .0; D = -2.26 \\ \cline{2-13} 
 & $\Delta$E$_{min}$ & -4.53 ± 0.78 & -2.03 ± 0.51 & -1.81 ± 0.12 & -1.51 ± 0.16 & -1.2 ± 0.47 & -1.8 ± 0.21 & \gradient{ .01} & \gradient{ .93} & \gradienteta{ .234} & \gradientetados{ .0} & Hypertensive \textgreater N; P = .006; D = 1.24 \\ \cline{2-13} 
 & $\Delta$A$_{max}$ & 2.6 ± 0.87 & 1.2 ± 0.44 & 2.26 ± 0.63 & 0.85 ± 0.13 & 0.23 ± 0.26 & 0.88 ± 0.12 & \gradient{ .004} & \gradient{ .007} & \gradienteta{ .261} & \gradientetados{ .123} &  \\ \cline{2-13} 
 & $\Delta$A$_{min}$ & -2.66 ± 0.94 & -2.33 ± 0.81 & -2.52 ± 0.77 & -1.2 ± 0.12 & -1.99 ± 0.66 & -2.15 ± 0.23 & \gradient{ .306} & \gradient{ .019} & \gradienteta{ .104} & \gradientetados{ .099} &  \\ \hline
\end{tabular}%
    }
\end{minipage}
\end{table}

\vspace{-1em}

\begin{table}[h!]
    \begin{minipage}{\textwidth}
    \caption{Average peak values and standard error of the mean for the flow rate parameters. The ANCOVA P-values and effect size ($\eta_{p}^{2}$) results are shown for the effect of pathology and age as a covariable. The P-values are color-coded, ranging from green (0.0) to red (0.2), indicating the level of statistical significance. Similarly, $\eta_{p}^{2}$ is represented using a color gradient, ranging from white (0.0) to blue (0.4) for pathology and white (0.0) to red (0.4) for age, indicating the strength of the association. Lastly, the posthoc results are shown for the pairwise comparisons between pathologies if they are found to be statistically significant ($\alpha<0.5$).}
    \label{tab:FlowPeaks}
    \small
    \resizebox{\textwidth}{!}{%
    \begin{tabular}{lllllllllllll}
\hline
&  & \multicolumn{6}{l}{Pathology} & \multicolumn{2}{l}{ANCOVA: P-value} & \multicolumn{2}{l}{ANCOVA: $\eta_{p}^{2}$} & Posthoc \\ \hline
 &  & Control & HCM & G1 & G2 & G2 - SAM & Hypertensive & Pathology & Age & Pathology & Age & Tukey and Cohen's D \\ \hline
E (ml/s)& \multirow{6}{*}{MV} & 350.71 ± 23.44 & 293.46 ± 42.31 & 290.24 ± 8.89 & 265.66 ± 28.6 & 224.81 ± 31.02 & 251.46 ± 16.56 & \gradient{ .644} & \gradient{ .003} & \gradienteta{ .052} & \gradientetados{ .135} &  \\ \cline{1-1} \cline{3-13} 
A (ml/s) &  & 191.29 ± 15.76 & 239.13 ± 26.78 & 285.92 ± 39.89 & 211.36 ± 34.64 & 184.53 ± 50.25 & 266.4 ± 11.41 & \gradient{ .037} & \gradient{ .009} & \gradienteta{ .17} & \gradientetados{ .105} &  \\ \cline{1-1} \cline{3-13} 
E$_{VOL}$ (ml) &  & 56.63 ± 4.39 & 51.72 ± 8.67 & 57.2 ± 9.83 & 49.06 ± 2.55 & 36.7 ± 9.19 & 39.56 ± 3.19 & \gradient{ .279} & \gradient{ .001} & \gradienteta{ .094} & \gradientetados{ .177} &  \\ \cline{1-1} \cline{3-13} 
A$_{VOL}$ (ml) &  & 18.18 ± 1.51 & 27.57 ± 3.34 & 32.58 ± 4.01 & 26.71 ± 3.73 & 28.84 ± 3.74 & 30.97 ± 1.64 & \gradient{ .038} & \gradient{ .007} & \gradienteta{ .169} & \gradientetados{ .112} & Hypertensive \textgreater N; P = .0; D = 1.63 \\ \cline{1-1} \cline{3-13} 
E/A &  & 2.15 ± 0.25 & 1.32 ± 0.22 & 1.08 ± 0.12 & 1.52 ± 0.35 & 2.3 ± 1.36 & 0.99 ± 0.08 & \gradient{ .014} & \gradient{ .015} & \gradienteta{ .202} & \gradientetados{ .091} & Hypertensive \textless N; P = .018; D = -1.49 \\ \cline{1-1} \cline{3-13} 
E$_{VOL}$/A$_{VOL}$ &  & 3.63 ± 0.41 & 2.13 ± 0.44 & 1.89 ± 0.43 & 2.2 ± 0.53 & 1.43 ± 0.53 & 1.4 ± 0.12 & \gradient{ .037} & \gradient{ .0} & \gradienteta{ .17} & \gradientetados{ .279} & Hypertensive \textless N; P = .0; D = -1.74 \\ \hline
\multirow{4}{*}{S (ml/s)} & RS  & 39.53 ± 3.67 & 45.26 ± 6.21 & 57.22 ± 4.87 & 42.15 ± 4.58 & 29.0 ± 7.6 & 48.24 ± 3.91 & \gradient{ .212} & \gradient{ .981} & \gradienteta{ .106} & \gradientetados{ .0} &  \\ \cline{2-13} 
 & RI  & 38.15 ± 3.28 & 31.26 ± 2.8 & 38.07 ± 10.19 & 36.23 ± 9.78 & 21.61 ± 7.39 & 40.5 ± 2.9 & \gradient{ .268} & \gradient{ .888} & \gradienteta{ .098} & \gradientetados{ .0} &  \\ \cline{2-13} 
 & LS  & 38.82 ± 3.44 & 39.36 ± 6.17 & 39.13 ± 1.66 & 32.86 ± 4.01 & 14.47 ± 4.57 & 42.79 ± 2.6 & \gradient{ .019} & \gradient{ .908} & \gradienteta{ .192} & \gradientetados{ .0} &  \\ \cline{2-13} 
 & LI  & 36.93 ± 3.46 & 26.93 ± 2.94 & 25.14 ± 5.4 & 23.56 ± 5.54 & 11.38 ± 4.96 & 33.87 ± 2.62 & \gradient{ .027} & \gradient{ .006} & \gradienteta{ .181} & \gradientetados{ .115} &  \\ \hline
\multirow{4}{*}{D (ml/s)} & RS  & 38.88 ± 3.49 & 35.74 ± 3.54 & 37.46 ± 4.01 & 33.3 ± 3.51 & 27.57 ± 7.45 & 33.69 ± 2.79 & \gradient{ .956} & \gradient{ .039} & \gradienteta{ .017} & \gradientetados{ .067} &  \\ \cline{2-13} 
 & RI & 37.52 ± 3.62 & 25.94 ± 3.76 & 25.07 ± 3.15 & 27.07 ± 4.96 & 22.7 ± 7.76 & 25.53 ± 2.2 & \gradient{ .634} & \gradient{ .025} & \gradienteta{ .053} & \gradientetados{ .079} &  \\ \cline{2-13} 
 & LS  & 37.57 ± 2.73 & 32.1 ± 6.09 & 24.23 ± 0.96 & 22.58 ± 2.88 & 13.23 ± 2.72 & 28.11 ± 1.9 & \gradient{ .24} & \gradient{ .029} & \gradienteta{ .101} & \gradientetados{ .075} &  \\ \cline{2-13} 
 & LI  & 34.89 ± 3.85 & 21.83 ± 3.75 & 16.83 ± 2.62 & 16.2 ± 3.08 & 14.08 ± 4.5 & 22.35 ± 2.23 & \gradient{ .282} & \gradient{ .001} & \gradienteta{ .094} & \gradientetados{ .158} &  \\ \hline
\multirow{4}{*}{S/D} & RS & 1.09 ± 0.08 & 1.3 ± 0.15 & 1.57 ± 0.15 & 1.27 ± 0.05 & 1.13 ± 0.31 & 1.49 ± 0.07 & \gradient{ .116} & \gradient{ .013} & \gradienteta{ .13} & \gradientetados{ .096} &  \\ \cline{2-13} 
 & RI & 1.11 ± 0.09 & 1.36 ± 0.17 & 1.43 ± 0.38 & 1.22 ± 0.18 & 0.98 ± 0.24 & 1.62 ± 0.08 & \gradient{ .016} & \gradient{ .003} & \gradienteta{ .2} & \gradientetados{ .138} &  \\ \cline{2-13} 
 & LS & 1.08 ± 0.08 & 1.35 ± 0.11 & 1.62 ± 0.09 & 1.47 ± 0.09 & 1.14 ± 0.45 & 1.58 ± 0.07 & \gradient{ .026} & \gradient{ .044} & \gradienteta{ .182} & \gradientetados{ .064} & Hypertensive \textgreater N; P = .01; D = 1.36 \\ \cline{2-13} 
 & LI & 1.15 ± 0.07 & 1.34 ± 0.1 & 1.45 ± 0.12 & 1.53 ± 0.33 & 0.91 ± 0.36 & 1.61 ± 0.09 & \gradient{ .018} & \gradient{ .055} & \gradienteta{ .194} & \gradientetados{ .058} &  \\ \hline
\multirow{4}{*}{Ar (ml/s)} & RS  & -5.78 ± 1.02 & 1.01 ± 2.12 & -10.15 ± 5.28 & -7.26 ± 6.87 & -1.41 ± 3.3 & 1.63 ± 1.75 & \gradient{ .018} & \gradient{ .384} & \gradienteta{ .194} & \gradientetados{ .012} &  \\ \cline{2-13} 
 & RI  & -9.28 ± 1.09 & -1.77 ± 1.49 & -8.62 ± 4.7 & 0.42 ± 3.24 & -3.31 ± 4.46 & -0.79 ± 1.2 & \gradient{ .002} & \gradient{ .823} & \gradienteta{ .265} & \gradientetados{ .001} & Hypertensive \textgreater N; P = .005; D = 1.49 \\ \cline{2-13} 
 & LS  & -7.25 ± 2.94 & -1.43 ± 2.16 & -12.04 ± 4.36 & -6.08 ± 3.96 & 1.03 ± 1.03 & -0.75 ± 1.44 & \gradient{ .078} & \gradient{ .574} & \gradienteta{ .145} & \gradientetados{ .005} &  \\ \cline{2-13} 
 & LI  & -9.45 ± 1.19 & -2.52 ± 1.7 & -10.79 ± 3.53 & -3.11 ± 2.18 & -2.88 ± 0.87 & -2.11 ± 0.99 & \gradient{ .001} & \gradient{ .575} & \gradienteta{ .269} & \gradientetados{ .005} & Hypertensive \textgreater N; P = .003; D = 1.4 \\ \hline
\end{tabular}%
    }
\end{minipage}
\end{table}

\clearpage

\section{PC-MRA: \texorpdfstring{$\gamma$} v values and single timestep segmentation}\label{pcmra_appendix}

\begin{figure}[h!]
\centering
\includegraphics[width=1\textwidth]{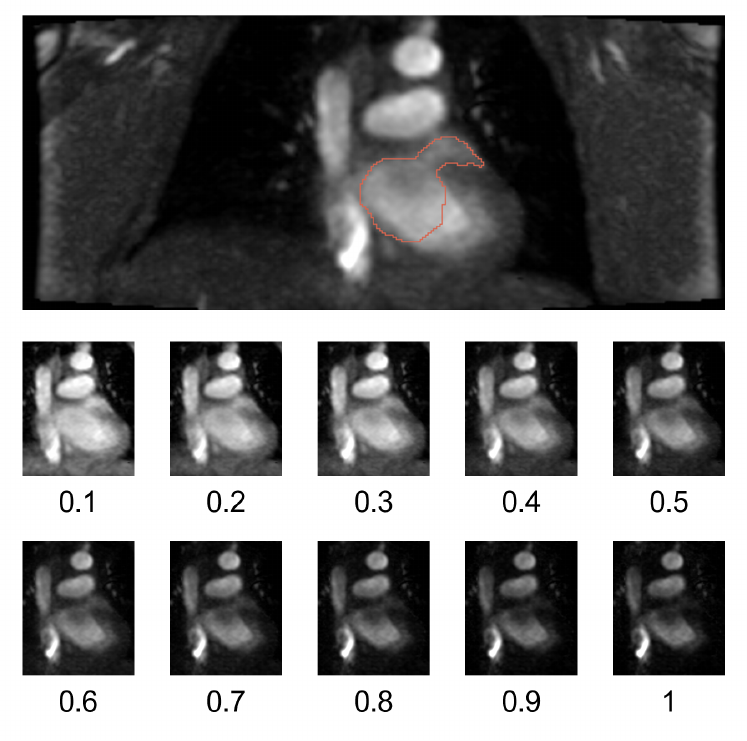}
\caption{Three-dimensional phase-contrast magnetic resonance angiograms (PC-MRA) were generated using the method described in Section \ref{sec: PC-MRA}. The coronal view of the chosen $\gamma$ value of 0.4 is shown at the top, with the outline of the left atrium segmentation in red. Below, the resulting 3D PC-MRA for 10 different $\gamma$ values are displayed. The lowest $\gamma$ values produce minimal contrast between soft tissues and blood, while low-velocity structures like the left atrial appendage (LAA) become nearly indiscernible at the highest $\gamma$ values. The intermediate $\gamma$ values strike a balance between blood pool and tissue contrast while highlighting low-velocity regions.  }
\label{Sup:Gamma}
\end{figure}

\clearpage

\begin{figure}[t!]
\centering
\includegraphics[width=\textwidth]{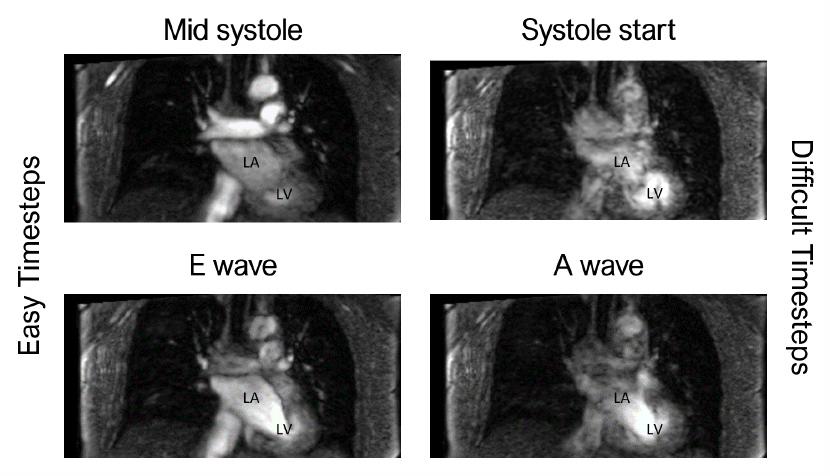}
\caption{Coronal view of the time-resolved four-dimensional phase-contrast magnetic resonance angiograms (PC-MRA) of a hypertrophic cardiomyopathy patient, generated using the equation proposed by \cite{Bustamante2017}. A $\gamma = 0.4$ was employed. During time steps when left atrial velocities are high, segmentation becomes feasible, as evidenced by the two examples in the left column, where the LA contour is clearly discernible. However, during periods of low velocity, such as the onset of the cardiac cycle or instances where high velocity is limited to specific areas (e.g., atrial contraction), distinguishing the LA from other cardiac structures might prove challenging or even infeasible.}
\label{Sup:Timesteps}
\end{figure}

\textcolor{white}{h}

\clearpage
\section{Segmentation results}\label{segfigures}

\begin{figure}[h!]
\centering
\includegraphics[width=\textwidth]{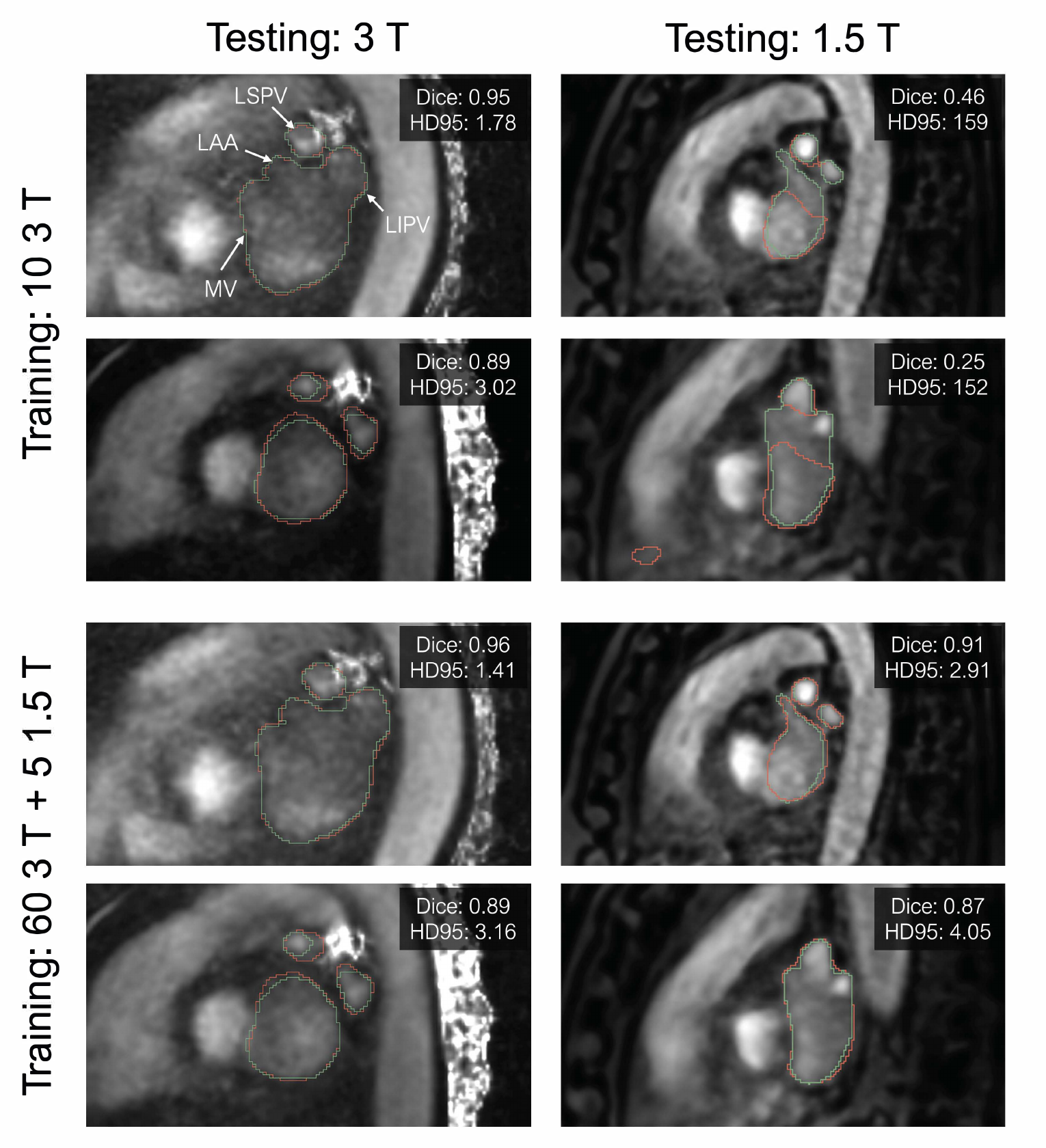}
\caption{Contours of the manually annotated ground truth segmentations (green) and nnU-Net predictions (red) overlaid over the time-averaged phase-contrast MR angiograms (PC-MRA) in the sagittal view. The first two rows show the best and worst-performing test cases from the first iteration of Experiment 1 (10 3T during training) using 1.5T and 3T datasets. The last two rows display the same cases after training with 60 3T and 5 1.5T cases (Experiment 2), showing significant improvement in 1.5T test cases, while the 3T cases slightly improve or remain consistent. Dice: Dice score, HD95: 95th percentile Hausdorff distance (mm), MV: Mitral valve, LAA: Left atrial appendage, LSPV: Left superior pulmonary vein, LIPV: Left inferior pulmonary vein.}
\label{Sup:SegResults}
\end{figure}

\end{document}